\documentclass[dvipsnames]{article} %
\usepackage{colm2024_conference}

\usepackage{booktabs}
\usepackage{enumitem}
\usepackage{wrapfig}
\usepackage{algorithm}
\usepackage{algpseudocode}
\usepackage{graphicx}
\usepackage[misc]{ifsym}
\usepackage{multicol}

\usepackage{microtype}
\usepackage{amsmath}
\usepackage{colortbl}
\usepackage[utf8]{inputenc}
\usepackage[T1]{fontenc}
\definecolor{lightgray}{rgb}{0.9,0.9,0.9}
\usepackage{caption}
\usepackage{subcaption}
\usepackage{setspace}
\usepackage{url}
\usepackage{multirow}
\usepackage{tabularx}
\usepackage{blindtext}
\usepackage{pgfplots}
\pgfplotsset{compat=1.18} 
\usepackage{tikz}
\usetikzlibrary{er,positioning,bayesnet}
\usepackage{makecell}
\usepackage{tipa}
\usepackage{siunitx}
\usepackage{nicefrac}
\usepackage{listings}
\usepackage[raster,skins, most]{tcolorbox} %
\usepackage{xltabular}
\usepackage{adjustbox}
\usepackage{xurl}
\usepackage{rotating}
\usepackage[normalem]{ulem}
\usepackage{caption}    
\usepackage{subcaption} 
\usepackage[subfigure]{tocloft}
\usepackage{fontawesome}
\usepackage{arydshln}

\useunder{\uline}{\ul}{}

\usepackage{amsmath,amsfonts,bm}
\usepackage{pifont}

\def\eqref#1{equation~\ref{#1}}

\def\1{\bm{1}}

\DeclareMathAlphabet{\mathsfit}{\encodingdefault}{\sfdefault}{m}{sl}
\SetMathAlphabet{\mathsfit}{bold}{\encodingdefault}{\sfdefault}{bx}{n}

\renewcommand{\ghlink}{https://github.com/Tongyi-MAI/MAI-UI}

\newcommand*\justify{%
  \fontdimen2\font=0.4em
  \fontdimen3\font=0.2em
  \fontdimen4\font=0.1em
  \fontdimen7\font=0.1em
  \hyphenchar\font=`\-
}

\renewcommand{\texttt}[1]{%
  \begingroup
  \ttfamily
  \begingroup\lccode`~=`/\lowercase{\endgroup\def~}{/\discretionary{}{}{}}%
  \begingroup\lccode`~=`[\lowercase{\endgroup\def~}{[\discretionary{}{}{}}%
  \begingroup\lccode`~=`.\lowercase{\endgroup\def~}{.\discretionary{}{}{}}%
  \catcode`/=\active\catcode`[=\active\catcode`.=\active
  \justify\scantokens{#1\noexpand}%
  \endgroup
}

\usepackage{makecell}
\usetikzlibrary{tikzmark}
\makeatletter
\newcommand*\myfontsize{%
  \@setfontsize\myfontsize{7}{8}%
}
\makeatother

\definecolor{lightblue}{RGB}{173,216,230}
\definecolor{uclablue}{RGB}{159, 195, 224}

\definecolor{uclagold}{RGB}{255, 240, 180}

\definecolor{aliceblue}{RGB}{255, 238, 241}

\definecolor{cadmiumgreen}{rgb}{0.0, 0.42, 0.24}

\definecolor{myred}{rgb}{0.7, 0.3, 0.0}
\definecolor{myblue}{rgb}{0.2, 0.3, 0.6}
\definecolor{babygreen}{rgb}{0.85, 0.97, 0.85}

\definecolor{purple1}{RGB}{126, 107, 196}
\definecolor{purple2}{RGB}{199, 158, 207}
\definecolor{purple3}{RGB}{214, 200, 255}
\definecolor{purple4}{RGB}{254, 240, 255}

\definecolor{deepblue}{RGB}{48, 58, 82}
\definecolor{light_purple}{RGB}{228, 228, 251}
\definecolor{deep_purple}{RGB}{234, 224, 240}
\definecolor{deep_purple2}{RGB}{102, 48, 226}

\newcommand{\symboletongyi}{\raisebox{0pt}{~\includegraphics[scale=0.012]{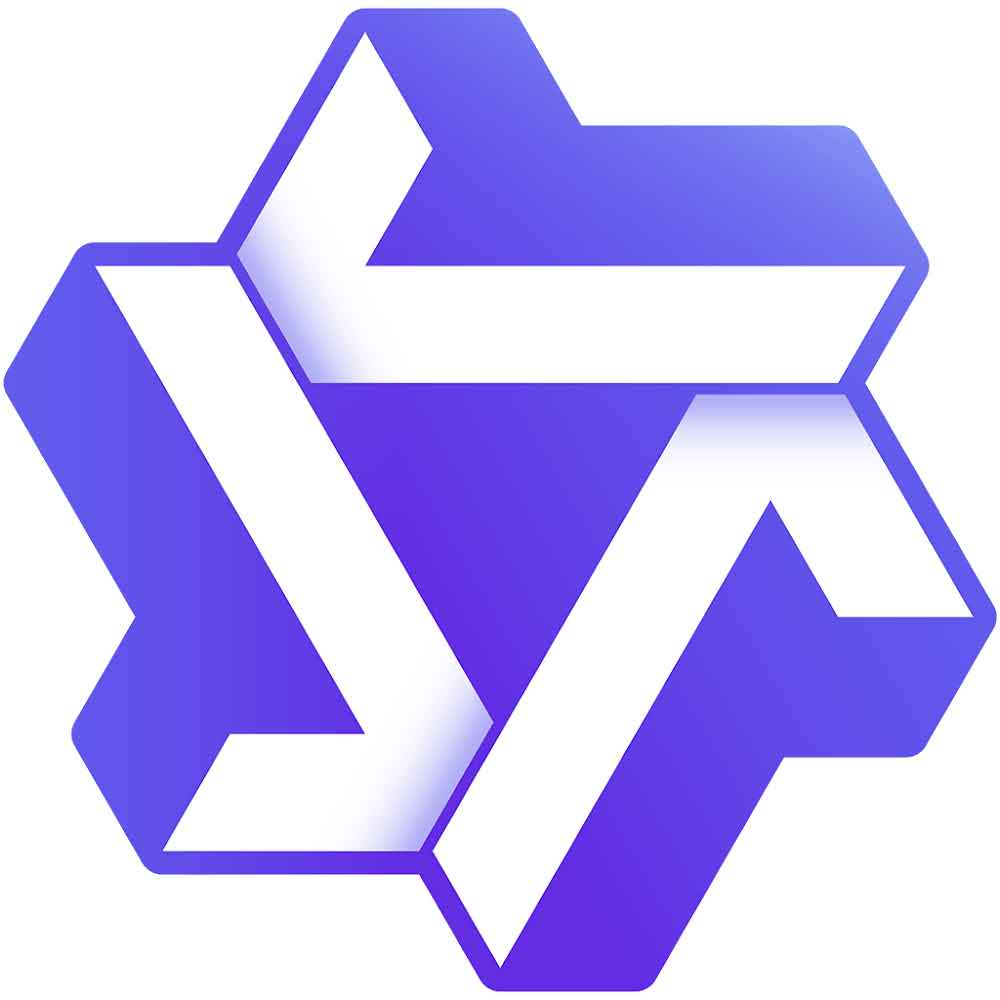}}~}

\definecolor{deepPurple}{HTML}{330066}

\definecolor{uclablue_old}{rgb}{0.15, 0.45, 0.68}
\hypersetup{
    breaklinks,
    citecolor=uclablue_old,
    colorlinks=true,
}
\usepackage[most]{tcolorbox} 
\newtcbtheorem{promptbox}{Prompt}{
    breakable,
    colframe=deepPurple!70,
}{box}

\newtcolorbox{mybox}[2][]
  {colback = black!5!white, colframe = black!75!black, fonttitle = \bfseries,
    colbacktitle = black!100!black, enhanced, before upper={\fontsize{8}{11}\obeyspaces\obeylines\selectfont}, fontupper=\selectfont,
    attach boxed title to top left={yshift=-2.2mm,xshift=4mm},
    title=#2,#1}
\newcommand{\mobileworld}{\textsc{MobileWorld}}

\newcommand{\modelname}{MAI-UI}
\title{
    \modelname{} Technical Report: \\Real-World Centric Foundation GUI Agents
}

\author{%
{Hanzhang Zhou$^{*}$, Xu Zhang$^{*}$, Panrong Tong$^{}$, Jianan Zhang$^{}$, Liangyu Chen$^{}$, Quyu Kong$^{}$\\Chenglin Cai$^{}$, Chen Liu$^{}$, Yue Wang$^{(\textrm{\Letter})}$, Jingren Zhou, Steven HOI$^{}$
}%
  \\[1em]               
  {\fontsize{10pt}{11pt}\selectfont          
Tongyi Lab\symboletongyi, Alibaba Group}\\
[0.5em]
}

\begin{document}

\maketitle

\begin{abstract}
\vspace{-0.5em}
The development of GUI agents could revolutionize the next generation of human-computer interaction. Motivated by this vision, we present \modelname{}, a family of foundation GUI agents spanning the full spectrum of sizes, including 2B, 8B, 32B, and 235B-A22B variants. We identify four key challenges to realistic deployment: the lack of native agent–user interaction, the limits of UI-only operation, the absence of a practical deployment architecture, and brittleness in dynamic environments. \modelname{} addresses these issues with a unified methodology: a self-evolving data pipeline that expands the navigation data to include user interaction and MCP tool calls, a native device–cloud collaboration system routes execution by task state, and an online RL framework with advanced optimizations to scale parallel environments and context length. \\
\vspace{+0.3em}
\modelname{} establishes new state-of-the-art across GUI grounding and mobile navigation. On grounding benchmarks, it reaches \textbf{73.5\%} on ScreenSpot-Pro, \textbf{91.3\%} on MMBench GUI L2, \textbf{70.9\%} on OSWorld-G, and \textbf{49.2\%} on UI-Vision, surpassing Gemini-3-Pro and Seed1.8 on ScreenSpot-Pro. On mobile GUI navigation, it sets a new SOTA of \textbf{76.7\%} on AndroidWorld, surpassing UI-Tars-2, Gemini-2.5-Pro and Seed1.8. On MobileWorld, \modelname{} obtains \textbf{41.7\%} success rate, significantly outperforming end-to-end GUI models and competitive with Gemini-3-Pro based agentic frameworks. Our online RL experiments show significant gains from scaling parallel environments from 32 to 512 (+5.2 points) and increasing environment step budget from 15 to 50 (+4.3 points).
Finally, the native device-cloud collaboration system improves on-device performance by \textbf{33\%}, reduces cloud model calls by over \textbf{40\%}, and preserves user privacy.

\end{abstract}

\begin{figure}[h]
\vspace{-1.1em}
    \centering  \includegraphics[width=0.97\linewidth]{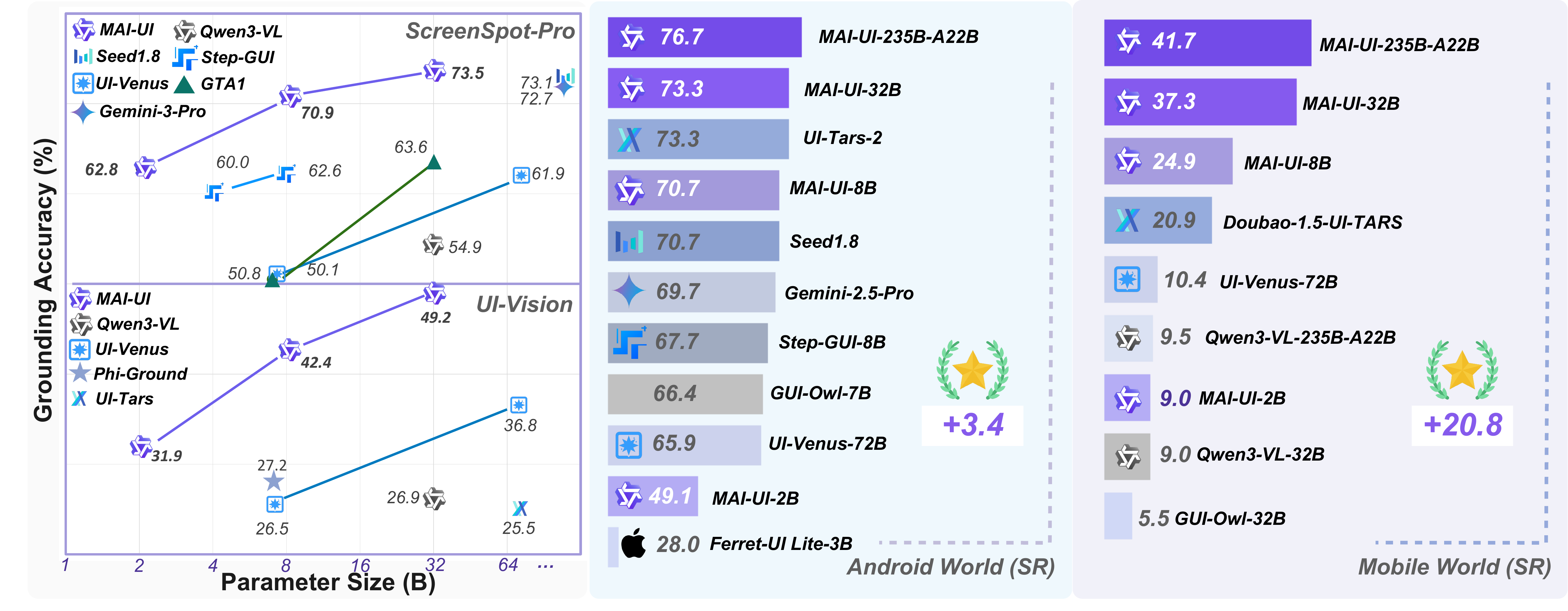}
    \vspace{-1.1em}
    \caption{\modelname{} achieves SOTA performance across GUI grounding and navigation benchmarks.}
    \label{fig:data_generation}
\end{figure}

\renewcommand{\thefootnote}{*}
\footnotetext{Lead contributor. All authors made core contributions.~~~\textrm{\Letter}~Corresponding author. \texttt{yue.w@alibaba-inc.com}}
\renewcommand{\thefootnote}{\arabic{footnote}}

\section{Introduction}
\label{sec:intro}
Graphical user interface (GUI) agent \citep{gui_agent_survey} is an agent system that can perceive, reason, and act within GUIs in response to natural language instructions. By translating high‑level user intents into concrete UI operations, GUI agents potentially represent a revolution in digital interaction, transforming human-computer interaction from the manual navigation of complex interfaces to goal‑oriented natural language control.  
Motivated by this vision, the community’s increasing efforts have accelerated progress on GUI agents, advancing UI perception, visual grounding, and GUI navigation \citep{uitars2, mobileagentv3, ui-venus}.
 
Despite rapid technological advances, today’s GUI agents remain insufficient for reliable, robust, and secure deployment in practice. We highlight several open challenges to close this gap. \textbf{(1) Agent-user interaction}: Existing systems are typically optimized for end-to-end execution, however, user instructions in real-world settings are often ambiguous or incomplete. To ensure alignment with user intent, the agent must proactively ask clarifying questions, collect missing details, and seek consent for sensitive operations. Effective agent–user interaction is therefore a critical yet often neglected capability.
\textbf{(2) Beyond UI-only action:} Relying solely on UI manipulation poses two issues: (i) the long, multi-step sequences of UI operations increase brittleness to per-step errors and amplify error propagation; and (ii) it limits the agent to tasks that are UI-reachable. Integrating external tools via the Model Context Protocol (MCP) provides structured shortcuts that compress long, fragile UI operation sequences into a few API calls and unlock tasks previously infeasible on mobile. For example, via MCP tools, a mobile agent can manipulate GitHub repositories, bringing traditionally desktop-only workflows to mobile. 
\textbf{(3) Native device–cloud collaboration capability:} Current GUI agents are typically categorized into lightweight, on-device variants or large models that can only be used as cloud services. However, cloud-only solutions introduce privacy risks, higher costs, and dependence on network connectivity, whereas on-device–only approaches are constrained by model capacity and capability. Consequently, foundation GUI agents lack native device–cloud collaboration capability for privacy-aware and cost-efficient routing and seamless handoff. \textbf{(4) Robustness to dynamic environments:} Agents trained on static, pre-collected trajectories often overfit to specific interface patterns and struggle in out-of-domain scenarios. In practice, real-world GUIs are highly dynamic: layouts vary across app versions and devices, and pop-ups or permission dialogs can appear unexpectedly. Without exposure to dynamic environments in training, agents generalize poorly and remain brittle to real-world unpredictability.

To enhance GUI agent capabilities and address these challenges for realistic deployment, we introduce \textbf{\modelname{}}, a foundational GUI agent for general GUI grounding and mobile navigation.
\begin{itemize}[leftmargin=*]
\vspace{-0.5em}
\item \textbf{Agent-user interaction and MCP augmentation.} To equip GUI agents with user interaction and MCP tool use capability, we introduce a self-evolving data pipeline that incorporates training data for general navigation as well as these two capabilities. The data pipeline iteratively updates both the model and the training corpus using three sources of data: rejection-sampled trajectories, manually-annotated trajectories, and automatic agent rollouts.  The action space is also extended to allow the agent to choose among UI manipulation, user engagement, and MCP tool use. 
\item \textbf{Device–cloud collaboration system.} For realistic deployment, \modelname{} introduces a pioneering native device-cloud collaborative system, which can dynamically select on-device or cloud execution based on task execution state and data sensitivity. The system consists of a local GUI agent that both act as a GUI agent and as trajectory monitor, a high-capacity cloud GUI agent, and a local unified trajectory memory that maintains consistent information exchange between local and cloud agents.
\item \textbf{Reinforcement learning in dynamic environments.} \modelname{} incorporates online reinforcement learning as a core training component, enabling improvement through interaction with dynamic environments. Our system-level optimizations scale to 500+ GUI environments for parallel rollouts. We further support asynchronous rollout and hybrid parallelism for training, enabling training on long-horizon GUI tasks with up to 50 interactive steps. This training stage yields improved GUI navigation accuracy and stronger robustness to real-world unpredictability.
\end{itemize}

\vspace{-0.5em}
\modelname{} includes a full-spectrum of sizes to meet real-world deployment constraints, ranging from efficient \textbf{2B} on device variants to mid size \textbf{8B} and \textbf{32B} models, and large scale \textbf{235B-A22B} models. Across sizes, our models achieve state-of-the-art performance against strong baselines at comparable scales. Notably, our \textbf{2B on-device model} achieves a relative improvement of \textbf{75.4\%} over Ferret-UI Lite \citep{Ferret-ui-lite}, our \textbf{mid size 8B and 32B models} surpasses GUI-Owl\citep{mobileagentv3}, Step-GUI \citep{step-gui}, and UI-Venus\citep{ui-venus}, and our \textbf{235B-A22B variants} outperform UI-Tars-2 \citep{uitars2}, Seed1.8 \citep{seed18}, and Gemini-2.5-Pro \citep{GeminiComputerUse} on AndroidWorld.

\modelname{} sets a new state of the art across diverse evaluation settings, including GUI grounding, offline and online mobile GUI navigation, and a realistic-oriented benchmark that incorporates MCP tool use and agent-user interaction, consistently outperforming prior works.
\begin{itemize}[leftmargin=*]
\vspace{-0.3em}
\item \textbf{Grounding.} Our model establishes new state of the art performance across five well known grounding benchmarks. Notably, \modelname{} achieves \textbf{67.9\%} on ScreenSpot-Pro (\textbf{73.5\%} with zoom-in), \textbf{91.3\%} on MMBench GUI L2, \textbf{70.9\%} on OSWorld-G (\textbf{75.0\%} on OSWorld-G-Refine), \textbf{47.1\%} on UI-Vision (\textbf{49.2\%} with zoom-in), and \textbf{96.5\%} on ScreenSpot-V2, substantially surpassing the strongest counterparts.
\item \textbf{Offline GUI navigation}. On Android Control, \modelname{} attains strong exact match accuracy. On GUI Odyssey, it achieves a success rate of \textbf{83.4\%}, outperforming prior methods.
\item \textbf{Online GUI navigation}. In challenging dynamic real-time environments, \modelname{}-235B-A22B achieves a new state of the art success rate of \textbf{76.7\%} on AndroidWorld, surpassing UI-Tars-2 \citep{uitars2}, Gemini-2.5-Pro \citep{GeminiComputerUse}, and Seed1.8 \citep{seed18}. Additionally, \modelname{}-32B, \modelname{}-8B, and \modelname{}-2B attain success rate of \textbf{73.3\%}, \textbf{70.7\%}, and \textbf{49.1\%} on AndroidWorld, respectively, all outperforming competitive models at comparable scales.
\item \textbf{Realistic-oriented evaluation.} To bridge the gap for more challenging and realistic online assessment, we adopt our MobileWorld benchmark \citep{mobile_world}, which comprises 
evaluations beyond pure GUI operations. \modelname{} obtains \textbf{41.7\%} success rate, surpassing end-to-end GUI model baselines by \textbf{+20.8} and competitive with agentic frameworks with GPT-5 or Gemini-3-Pro as planners. Furthermore, on tasks that require agent-user interaction and MCP tool use, \modelname{} gains a success rate of \textbf{37.5\%} and \textbf{51.1\%}, respectively, representing an absolute increase of \textbf{+32.1} and \textbf{+18.7}.
\end{itemize}

\section{\modelname{}}
\begin{figure}
    \centering  \includegraphics[width=0.95\linewidth]{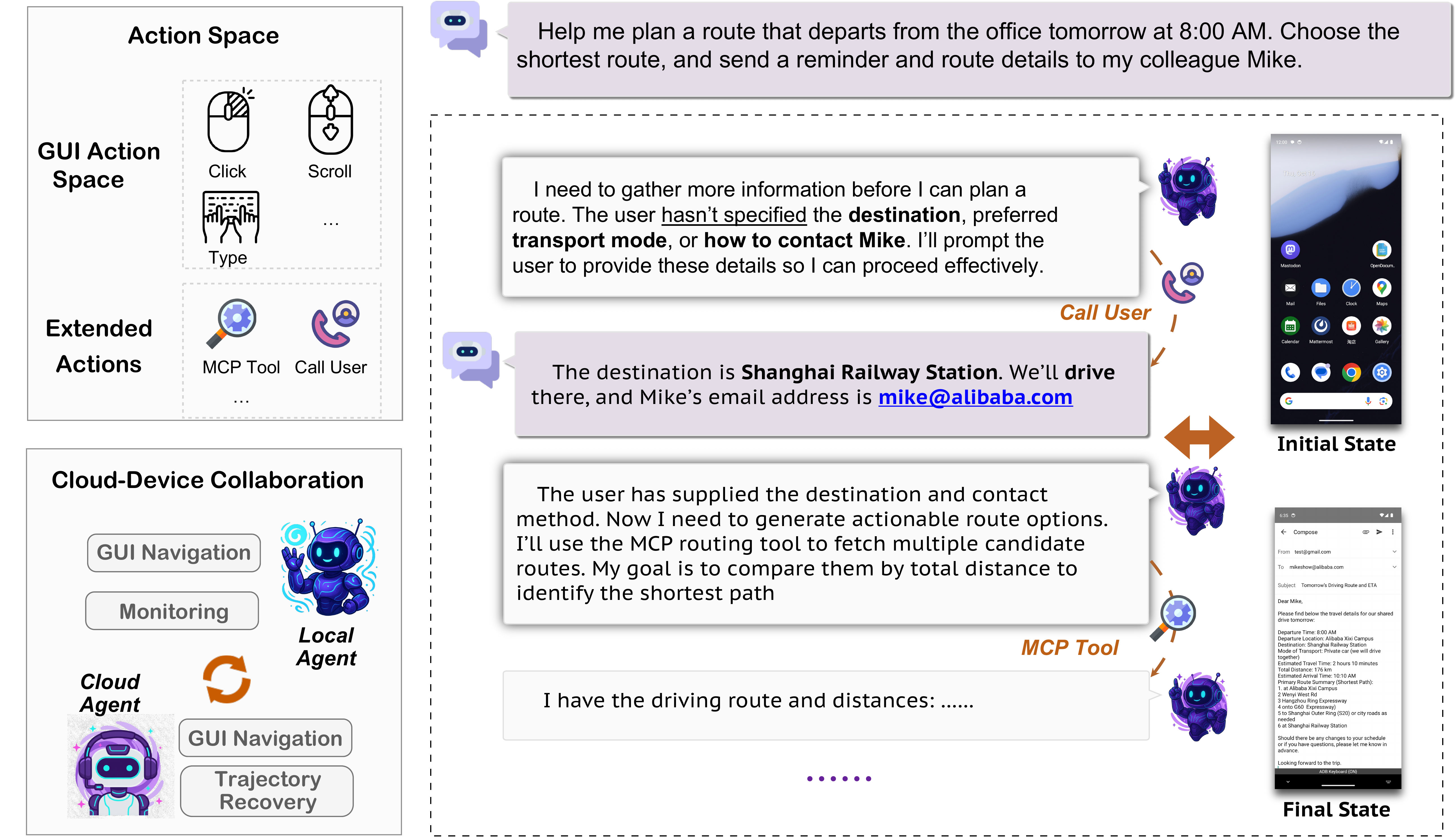}
    \caption{Example trajectory of \modelname{}. \modelname{} completes GUI agent tasks via both UI operations and extended actions, including agent-user interaction and MCP tool use, and integrates a native device-cloud collaboration system.}
    \label{fig:demo}
    \vspace{-0.5em}
\end{figure}

\vspace{-0.5em}
This section details the methodology of \modelname{}. Our approach integrates (i) a training paradigm for GUI grounding, (ii) a self-evolving trajectory data pipeline, (iii) training for agent–user interaction and MCP tool augmentation, (iv) online reinforcement learning, and (v) a native device–cloud collaboration system. Figure \ref{fig:demo} illustrates a demo trajectory of \modelname{}. 
\subsection{System Overview}
\vspace{-0.5em}
 We present a system overview of our model, covering task formulation, the action space, and the model architecture.
\label{sec:system overview}
\vspace{-0.5em}
\subsubsection{Task Formulation}
\modelname{} cover tasks of two categories: general GUI grounding and mobile GUI navigation.

\textbf{Grounding Task.}  
GUI grounding aims to localize the UI element corresponding to a natural language instruction on a graphical interface~\citep{wang2024ponderpressadvancing}. Formally, given a GUI screenshot and a natural language instruction $\mathcal{I}$, the GUI agent model predicts a coordinate point $\mathcal{P} = (x, y)$ that indicates the location of the target UI element.

\textbf{Navigation Task.} Mobile navigation task can be formulated as a {Partially Observable Markov Decision Process (POMDP)} $(\mathcal{S}, \mathcal{O}, \mathcal{A}, \mathcal{T})$~\citep{uitars}, where: $\mathcal{S}$ denotes the state space capturing the underlying environment; $\mathcal{O}$ represents the observation space, which consists of a natural language instruction $\mathcal{I}$ along with one or more screenshots; $\mathcal{A}$ defines the action space comprising standard mobile UI operations (e.g., click, swipe or typing); $\mathcal{T}: \mathcal{S} \times \mathcal{A} \rightarrow \mathcal{S}$ specifies the state transition of the environment.
At time step $t$, the agent predicts the next action as as $a_t = \pi(\mathcal{I}, o_t, h_t)$, where $\mathcal{I}$ is the natural language instruction, $o_t \in \mathcal{O}$ is the current observation, and $h_t = (a_1, o_1, \dots, a_{t-1}, o_{t-1})$ denotes the history context of previous actions and observations.

\subsubsection{Foundation GUI Agent for Mobile Use}
We introduce \modelname{}, a foundation GUI agent with strong GUI grounding and mobile navigation capabilities. It supports effective agent–user interaction for clarification and integrates MCP‑based tool augmentation for API use, and features a native device–cloud collaboration system for practical deployment.
\modelname{} offers a broad range of model sizes, each specifically trained to deliver robust grounding and navigation capabilities.
\begin{table}[h]
\centering
\caption{Action Space in \modelname{}.}
\label{tab:action_space}
\begin{tabular}{ll}
\hline
\textbf{Action} & \textbf{Definition} \\
\hline
\texttt{click} & Clicks at coordinates (x, y). \\
\texttt{long\_press} & Long presses at coordinates (x, y). \\
type & Types the specified text content. \\
\texttt{swipe} & Swipes in the given direction (up/down/left/right) at optional coordinates. \\
\texttt{drag} & Drags from start coordinates (x1, y1) to end coordinates (x2, y2). \\
\texttt{system\_button} & Presses a system button (back, home, menu, or enter). \\
\texttt{wait} & Pauses for a brief moment. \\
\texttt{terminate} & Marks the task as complete with status (success or fail). \\
\texttt{answer} & Provides an answer with specified text. \\
\texttt{ask\_user} & Requests user intervention with specified text. \\
\texttt{mcp\_call} &Provides MCP tool name and corresponding arguments. \\
\hline
\end{tabular}
\vspace{-0.5em}
\end{table}
\vspace{-0.7em}
\paragraph{Action Space.} \modelname{} provides a comprehensive action space for mobile GUI control, including \texttt{click}, \texttt{swipe}, \texttt{type}, \texttt{wait}, etc. 
By leveraging these actions, \modelname{} can interact with the mobile device through its graphical user interface to complete a wide range of tasks. 
We also include an \texttt{answer} action to directly respond to user queries in question-answering scenarios.\\
To better support real-world scenarios, we integrate two specialized actions that extend the agent beyond pure GUI operation through active user interaction and MCP tool use.
The \texttt{ask\_user} action allows the agent to request clarification when the instruction is vague or underspecified. 
The \texttt{mcp\_call} action enables the agent to leverage external MCP tools rather than relying solely on GUI operations.
The full action space is shown in Table~\ref{tab:action_space}.
\vspace{-0.7em}
\paragraph{Model Architecture.} We employ Qwen3-VL \citep{Qwen3-VL} as the backbone model. The \modelname{} family spans a full spectrum of sizes, including 2B, 8B, 32B, and 235B-A22B variants, enabling the deployment tailored to hardware constraints and performance requirements. 
Each model is jointly trained on GUI grounding, perception, and mobile-use navigation data using supervised fine-tuning and reinforcement learning. In addition, \modelname{} integrates a native device–cloud collaboration system that routes computation by task state and data sensitivity.

\subsection{GUI Grounding \& Perception}
\label{sec:grounding_perception}
GUI grounding and perception are foundational capabilities of GUI agents, enabling the agent to understand screen layouts and to localize the correct UI elements from natural-language instructions. This section presents our data pipeline and training methodology for building these capabilities. 
\subsubsection{Data Pipeline}

\begin{figure}
    \centering
    \includegraphics[width=0.95\linewidth]{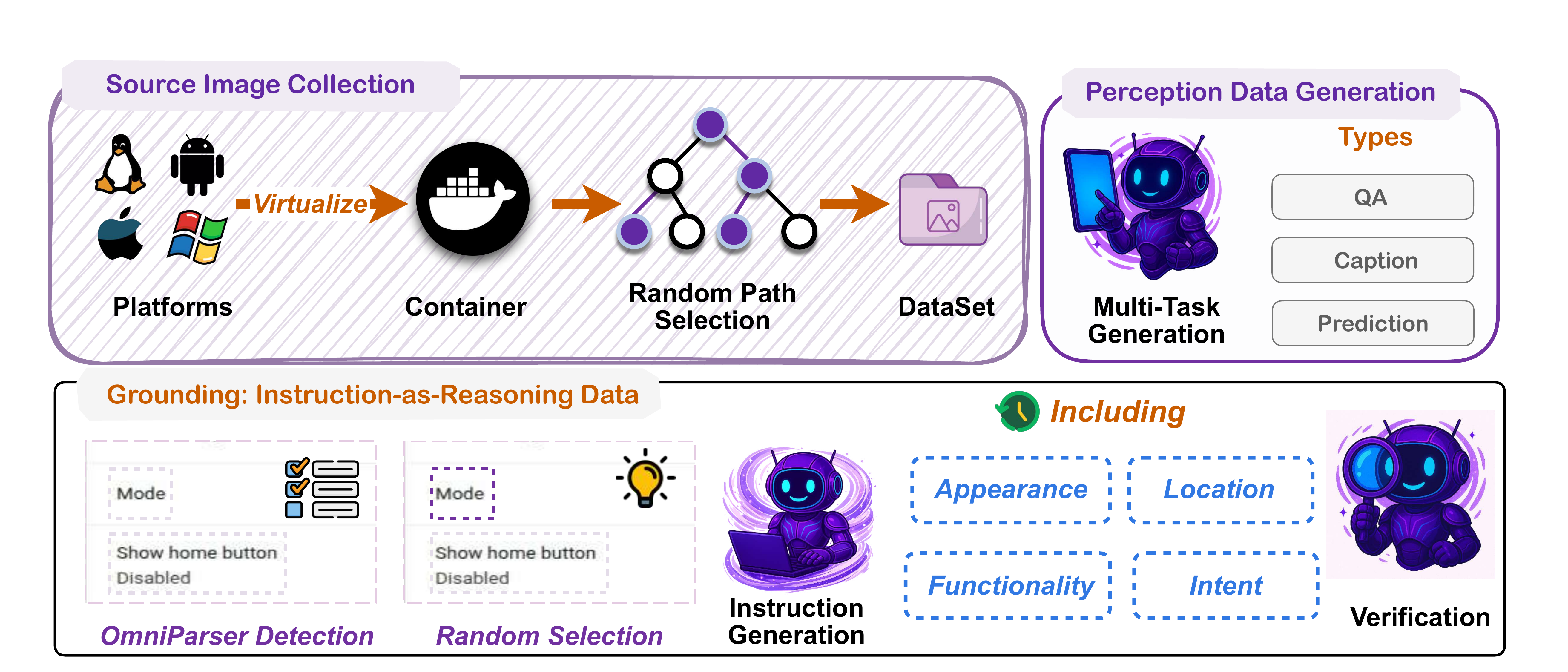}
    \caption{Overview of grounding and perception data pipeline}
    \label{fig:grounding_perception_data_pipeline}
    \vspace{-1.0em}
\end{figure}

As shown in ~\autoref{fig:grounding_perception_data_pipeline},  in addition to open-source datasets, our pipeline collects screenshots from real GUI environments to build a robust GUI Agent. This process results in multi-task perception data and multi-perspective grounding data.
\vspace{-1em}
\paragraph{Data Collection.} To gather diverse GUI data from real-world scenarios, we not only use open-source datasets such as JEDI \citep{jedi} and OS-Altas \citep{osaltas_and_screenspot_v2}, we also virtualize operating systems in containerized environments. We employed an MLLM-guided exploration strategy to navigate these environments. In every step, we ask MLLMs to identify valid actions from the current state. This procedure continuously changes the interface state and produces new screenshots. Finally, we use the a11y tree or OmniParser V2~\citep{omniparser} to localize UI elements precisely.
\vspace{-1em}
\paragraph{Perception Data Generation.} To facilitate the diversity in the training data, for each screenshot, we randomly select one to three UI elements as inputs and prompt MLLMs to generate a variety of tasks based on these elements, including question answering, captioning, and state prediction. These diverse tasks allow the model to develop a comprehensive understanding of the interface, specifically enhancing its capabilities in semantic understanding, relation understanding, and layout understanding.
\vspace{-1em}
\paragraph{Grounding Data Generation.}
As demonstrated in our prior work in UI-Ins~\citep{uiins}, instruction diversity and correctness are critical for GUI grounding but are often neglected:
\vspace{-1em}
\begin{itemize}[leftmargin=*]
\item Instruction quality: Approximately 23.3\% of instructions in open-source grounding datasets exhibit quality issues, which can actively harm model performance during training.
\item Instruction diversity: Performance improves significantly when training on diverse instruction perspectives, and improves further when the model learns to select appropriate instruction perspective in different scenarios.
\end{itemize}
\vspace{-0.5em}
Following our Instruction-as-Reasoning paradigm introduced in UI-Ins \citep{uiins}, we ask MLLMs to create instructions from four different human-like perspectives, including appearance, function, location and intent. This design is motivated by human UI grounding behavior: people strategically switch among instruction perspectives, selecting the most informative one for the task to support effective reasoning. We use these instructions both as inputs and as explicit reasoning pathways, thereby injecting structured reasoning into the model. This training method is introduced in the section below.
\subsubsection{Training Paradigm}
\paragraph{Algorithm.} To build a foundation model with strong grounding capability, we follow our Instruction-as-Reasoning paradigm in UI-Ins~\citep{uiins}. Using the grounding data described above, we first perform supervised fine-tuning (SFT) to instill instruction-as-reasoning capability: utilizing diverse instruction perspectives as explicit analytical reasoning before predicting the coordinates. To encourage dynamic, context-aware selection of the appropriate reasoning perspective across different scenarios, we then conduct a reinforcement learning (RL) stage using the GRPO algorithm.
\vspace{-0.5em}
\paragraph{Reward.} For GUI grounding, we use a combination of a format reward and a point-in-box reward. In our experiments, dense reward formulations yield similar performance, so we adopt this simple and effective scheme.
\vspace{-0.25em}
\begin{itemize}[leftmargin=*]
\item \textbf{Format Reward.} We use format reward $R_f$ verify that the model's response is in a valid format. Specifically, the thinking content and the final answer must be contained in their correct tags, and the coordinates must be extractable for the point-in-box evaluation.
\item \textbf{Point-in-Box Reward.} We utilize a direct point-in-box reward to measure correctness during training.  A prediction is considered correct if the predicted coordinate point $p = (x_p, y_p)$ falls within the ground-truth bounding box $b = (x_l, y_l, x_r, y_r)$, where the $(x_l,y_l)$ denotes the top-left corner and $(x_r,y_r)$ represents the bottom-right corner.
\begin{equation}
    R_{acc} = \begin{cases} 
        1 & \text{if } x_l \le x_p \le x_r \text{ and } y_l \le y_p \le y_r, \\
        0 & \text{otherwise.}
    \end{cases}
\end{equation}
\end{itemize}
\paragraph{Zoom-In Strategy.}
During inference, we introduce an optional zoom-in strategy for complex and high-resolution GUI scenarios. In the first pass, the model predicts a coarse coordinate. We then crop a window centered on this point, with width and height equal to half of the original image dimensions, and resize the crop back to the original resolution. In the second pass, the model refines the prediction by outputting the precise coordinate on the zoomed region.

\subsection{Mobile GUI Navigation}
\label{sec:mobile_use_navigation}
Beyond GUI perception and grounding in general setting, \modelname{} excels in mobile GUI navigation tasks. The training pipeline of GUI navigation consists of two main stages: supervised fine-tuning (SFT) and online reinforcement learning (RL). 
In the first stage, we build a self-evolving data pipeline to collect and synthesize diverse multi-step trajectories, and train the model with these trajectories to obtain strong navigation capability. 
In the second stage, we enhance the model’s generalization to real-world scenarios through online reinforcement learning in dynamic environments. The following sections provide detailed descriptions of both components. 
\begin{figure}
    \centering  \includegraphics[width=0.85\linewidth]{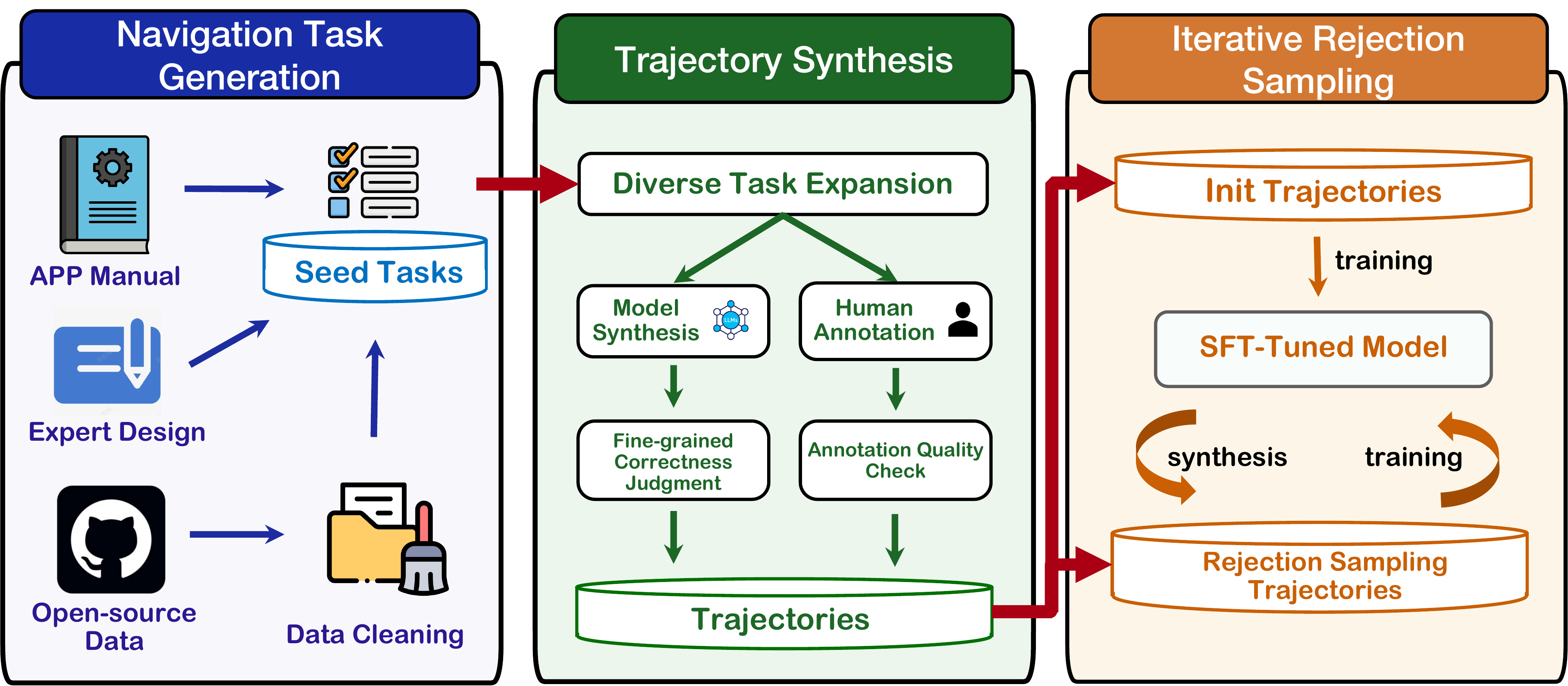}
    \caption{Overview of the self-evolving data pipeline for trajectory synthesis. The pipeline comprises task generation, trajectory construction via human annotation and autonomous agent rollouts, and iterative rejection sampling that jointly evolve the model and the training corpus.}
    \label{fig:data_generation}
\end{figure}
\subsubsection{Supervised Fine-Tuning}
\modelname{} employs a self-evolving SFT data pipeline (overview in Figure \ref{fig:data_generation}), comprising three key components: {Navigation Task Generation}, {Trajectory Synthesis}, and {Iterative Rejection Sampling}.  
In the \textbf{Navigation Task Generation} stage, we leverage multiple sources (APP manuals, expert-designed tasks, and open-source data) to construct high-quality seed tasks. 
In the \textbf{Trajectory Synthesis} stage, we first expand the seed tasks, then combine model-based synthesis and human annotation to generate diverse trajectories. The generated trajectories undergo two quality control steps: fine-grained correctness judgment via automated evaluation and annotation quality checks by human reviewers. This ensures both diversity and quality of the synthesized trajectories.
In the final stage, \textbf{Iterative Rejection Sampling}, we initialize training with a set of Stage 2 trajectories as cold-start data to obtain an initial SFT model. We then alternate between fine-tuning the model and deploying the updated policy to rollout new trajectories. Newly generated trajectories are filtered via rejection sampling, retaining only high-quality examples aligned with the model’s evolving capabilities. In parallel, we continually inject new trajectories from Stage 2 to broaden coverage and raise the performance ceiling. This closed loop of training and data synthesis makes both the model and the training corpus self-evolving.
\vspace{-0.5em}
\paragraph{Navigation Task Generation.}
In this stage, navigation task instructions are derived from three distinct sources: 
(1) \emph{application manuals}, from which common usage scenarios are parsed and distilled into intent-level task descriptions; 
(2) \emph{expert-designed tasks}, where human annotators formulate realistic and diverse mobile navigation goals aligned with commonly used scenarios; and 
(3) \emph{open-source datasets}, which are filtered by task complexity and reachability.
This multi-source strategy expands both task diversity and scale, while rigorous source selection and filtering ensures data quality.
\vspace{-0.5em}
\paragraph{Trajectory Synthesis.}
In the second stage, we first expand the set of tasks to increase diversity, and then generate trajectory data through parallel pipelines that leverage both model-based rollouts and human annotation.
\vspace{-0.5em}
\begin{itemize}[leftmargin=*]
    \item \textbf{Seed task expansion.} Starting from seed tasks, we prompt a multimodal large language model (MLLM) to generate a variety of novel tasks.
We categorize this diversity into two levels:
\textbf{L1} adjusts critical parameters of the original task goal, and typical parameters include date/time ranges, numeric thresholds, sorting/filter criteria, etc;
\textbf{L2} replaces the core objects involved in the task while remaining constrained to the same scenario and set of applications.

\item \textbf{Model synthesis and human annotation.} Given the expanded set of tasks, we generate execution trajectories via two parallel pipelines: (1) \textit{human annotation}: annotators manually perform tasks on an Android emulator and record both screenshots and ground-truth action sequences at each step. (2) \textit{model-based synthesis}: since human annotation is time‑consuming and costly, we also adopt multiple GUI agents to automatically produce valid action sequences for navigation tasks. Notably, for a given task goal, multiple valid execution paths often exist. By combining these complementary sources, we significantly broaden trajectory coverage and enhance dataset robustness.

\item \textbf{Fine-grained correctness judgment and annotation quality validation.}  
After generating diverse trajectories from both model synthesis and human annotation, we perform quality assessment through two independent validation pipelines:  

\begin{enumerate}[leftmargin=*]
    \item \textbf{Manual quality check:} All human-annotated trajectories are reviewed by a second annotator who verifies alignment between the action sequence, screenshots, and the original task goal. Inconsistent or ambiguous demonstrations are either corrected or discarded.

    \item \textbf{Fine-grained correctness judgment.} Trajectories produced by GUI agents rollouts are examined by an MLLM-as-a-judge module~\citep{gu2025surveyllmasajudge}. 
    The judge analyzes the task instruction, action history, and screenshots to assess correctness at both trajectory and step levels.
    Because some expanded tasks are infeasible and rollout models can struggle in complex tasks, many generated trajectories fail to fully complete the intended goal. However, failed trajectories often contain a substantial prefix of correct actions, with errors typically occurring only at intermediate or later steps. Recognizing that not all steps in a failed trajectory are erroneous, we adopt a fine‑grained judging approach to identify and retain useful sub‑trajectories. \\The evaluation comprises two components: (1) \textit{Overall Trajectory Judgment:} The MLLM-as-a-judge assesses end-to-end success, prioritizing visual evidence from screenshots over textual claims generated by the GUI agent; (2) \textit{Erroneous Trajectory Reuse:} For failed trajectories, the judge identifies the longest prefix of correct actions before the first deviation. 
    This enables reuse of failed trajectories, reducing data waste and enabling the model to learn from partial successes.
\end{enumerate}
\end{itemize}

\paragraph{Iterative Rejection Sampling.}
We adopt an iterative self-improvement loop that progressively refines both the model and the trajectory data distribution.  
Let \(\mathcal{M}^{(t)}\) denote our model after the \(t\)-th round of fine-tuning, and let \(\mathcal{I}_{\text{expansion}}\) be the set of diverse task instructions from \textit{Diverse Task Expansion}.  
In round \(t+1\), we use \(\mathcal{M}^{(t)}\) as the rollout policy to generate new trajectories on \(\mathcal{I}_{\text{expansion}}\):
\begin{equation}
    \mathcal{D}_{\text{RS}}^{(t+1)} = \big\{ Rollout (\mathcal{M}^{(t)}, i) \mid i \in \mathcal{I}_{\text{expansion}} \big\},
\end{equation}
where each rollout is filtered through the \textit{fine-grained correctness judgment} module to retain only high-quality or partially correct segments.  
The training set for the next iteration is then constructed by mixing the newly generated rejection sampling data with novel trajectories synthesized from the Trajectory Synthesis stage:
\begin{equation}
    \mathcal{D}^{(t+1)} =  \cdot \mathcal{D}_{\text{RS}}^{(t+1)} \cup   \cdot \mathcal{D}_{\text{synthesis}},
\end{equation}
where \(\mathcal{D}_{\text{synthesis}}\) denotes the trajectories generated from the human annotation and the agent rollouts.  
The model \(\mathcal{M}^{(t+1)}\) is then fine-tuned on \(\mathcal{D}^{(t+1)}\), completing the iteration. Rejection-sampled data helps close the gap between pass@1 and pass@N, while novel trajectories introduced in each iteration continually raise the pass@N performance ceiling.
This process encourages the data distribution to gradually align with the model’s evolving capabilities.
\subsubsection{Enabling Agent-User Interaction and MCP Tool Use}
To teach the model to interact with users and use MCP tools, we augment our self‑evolving data pipeline with trajectories that explicitly cover agent-user interaction and MCP augmentation. These trajectories are mixed into SFT, enabling the model to learn when to elicit missing information from the user and when to use MCP tools to complete tasks efficiently.
\vspace{-0.5em}
\paragraph{Agent–user interaction.} We construct tasks with deliberately omitted critical information. When the annotation/rollout reaches to a step that requires missing information, it issues an \texttt{ask\_user} action. The query is then routed to a synthetic user agent implemented with a standard LLM that is conditioned on hidden context containing the missing information. The user agent returns concise, context‑appropriate replies, including clarifications, corrections, or refusals when applicable. We log the query–response pairs in the history and continue the annotation/rollout so the trajectory can incorporate the returned information to complete the task. We generate both single‑turn and multi‑turn interactions.
\vspace{-0.5em}
\paragraph{MCP augmentation.} We design tasks that require or benefit from external MCP tools (e.g., Amap, Github, Stockstart). During trajectory annotation or rollout, the annotator/agent can issue \texttt{mcp\_call} with a tool name and arguments. The MCP server executes the call and returns structured outputs. We record tool schemas, arguments, results, and the subsequent UI actions, and we keep trajectories that demonstrate correct MCP tool selection.

\subsubsection{Online Reinforcement Learning}

\begin{figure}
    \centering  \includegraphics[width=0.95\linewidth]{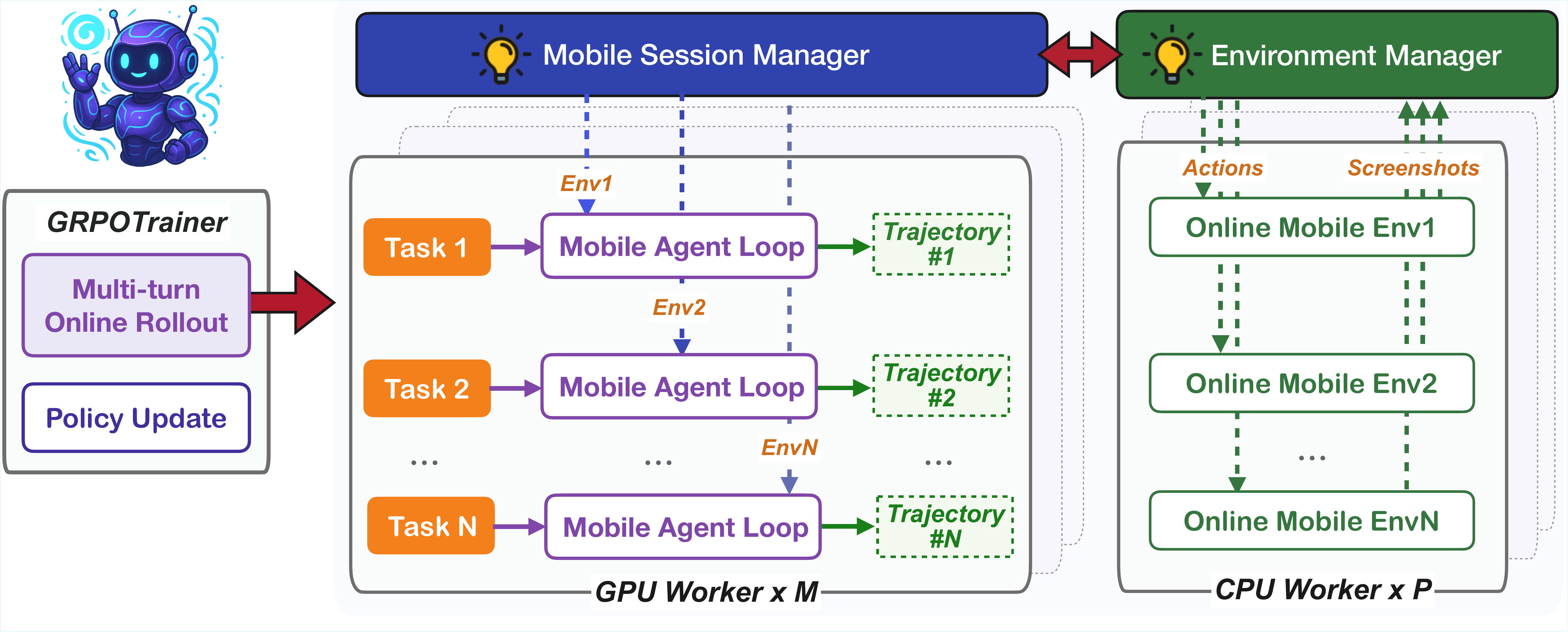}
    \caption{Overview of the agentic reinforcement learning framework. The framework alternates between rollout phases where the latest policy interacts with online mobile environments to generate trajectories and training phases that progressively improve the policy using trajectory-level rewards.}
    \label{fig: rl_overview}
\end{figure}

To enhance the model's reliability in long-horizon tasks and dynamic real-world environments, we employ an agentic reinforcement learning (RL) framework integrated with an online GUI environment, as illustrated in Figure \ref{fig: rl_overview}. This framework operates through two alternating phases that drive iterative improvement: (1) during the \textbf{rollout phase}, the model executes multi-turn interactions with the GUI environment to complete tasks and collect full execution trajectories; (2) in the \textbf{training phase}, the model performs end-to-end policy updates using trajectory-level rewards. Through this iterative process,  each improved policy generates higher-quality rollouts for subsequent training, progressively strengthening the model's robustness and generalization capability.
\paragraph{Scalable GUI Environment.} A critical bottleneck in scaling agentic RL lies in efficiently scaling stateful environments. Unlike stateless environments for mathematical reasoning or code generation, GUI environments are inherently stateful and resource-intensive, requiring each rollout to operate within an isolated instance. This constraint motivates our use of virtualized environments rather than physical devices.

Inspired by an experimental feature in AndroidWorld \citep{android_world}, we built a containerized solution that encapsulates the entire GUI environment within a Docker image, comprising a rooted Android Virtual Device (AVD), self-hosted backend services, and a dedicated REST API server for orchestration. This solution is carefully designed to ensure the following three features:
\vspace{-0.5em}
\begin{itemize}[leftmargin=*]
    \item \textbf{Consistency.} The unified containerization eliminates external dependencies and guarantees behavioral consistency across heterogeneous host systems.
    \item \textbf{Generalizability.} To build a general environment for mobile use, we integrated over 35 applications, encompassing system utilities and open-source software such as Mattermost (enterprise communication), Mastodon (social media), and Mall4Uni (e-commerce). Self-hosting these applications provides full backend access, enabling precise manipulation of initial task states and deterministic verification of execution outcomes.
    \item \textbf{RL-native Design.} We employ an AVD snapshot mechanism for reproducible task initialization and expose standard RL primitives ($\mathtt{reset}$, $\mathtt{step}$, $\mathtt{get\_observation}$, $\mathtt{evaluate}$, and $\mathtt{close}$) through the containerized API server, enabling parallel deployment of emulator instances.
\end{itemize}

To further scale environments across distributed infrastructure, we introduce a centralized Environment Manager that coordinates container instances across multiple physical machines. This architecture serves three critical functions: (1) \textit{efficient resource utilization}: through automatic container reuse, environments are reset and reassigned upon rollout completion instead of being destroyed, (2) \textit{cross-machine orchestration}: the Manager exposes a unified REST API that provides transparent access to distributed resources across heterogeneous hosts, and (3) \textit{fault tolerance}: we introduce automatic detection and recovery mechanisms to handle container failures, with failover protocols that seamlessly replace compromised instances from a standby pool. By coordinating just 10 standard Alibaba Cloud ECS servers (ecs.ebmg5s.24xlarge), the Manager supports up to \textbf{512} concurrent environment instances for parallel rollout execution, while maintaining high availability essential for continuous online RL training.

\begin{figure}
    \centering  \includegraphics[width=0.95\linewidth]{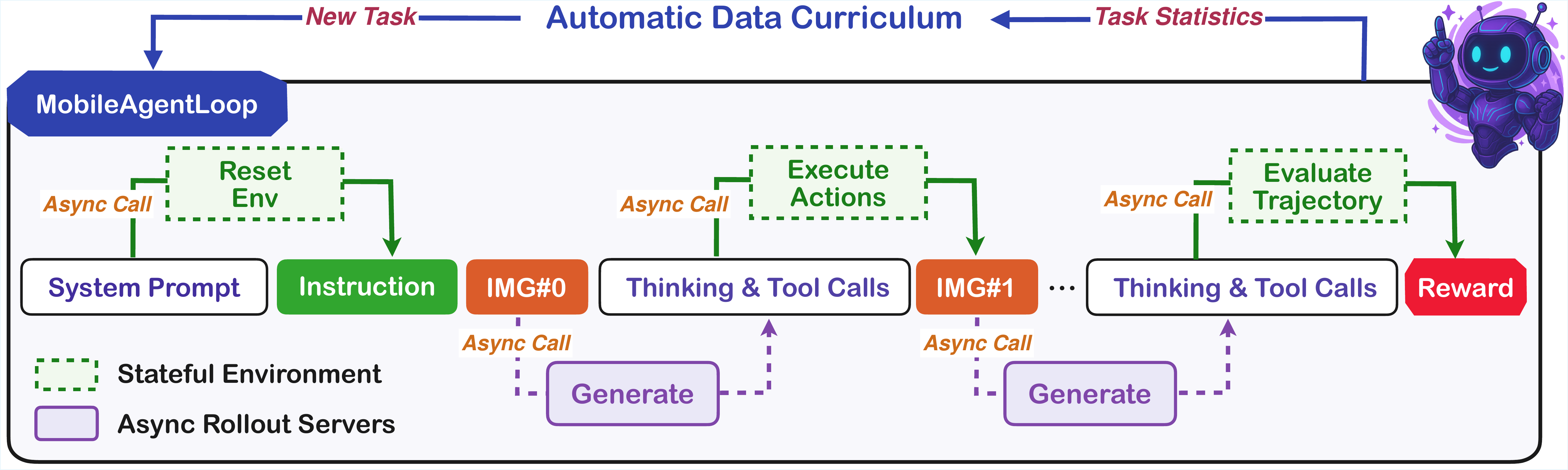}
    \caption{Detailed rollout process within the MobileAgentLoop. The agent loop asynchronously calls inference servers to generate actions and executes them in stateful environments across multiple turns, with hybrid verifiers evaluating complete trajectories to produce final rewards.}
    \label{fig: rl_agentloop}
\end{figure}

\vspace{-0.5em}
\paragraph{Long-Horizon RL.} Training RL agents for long-horizon tasks faces two interconnected challenges: traditional synchronous rollout pipelines become inefficient bottlenecks due to extensive multi-turn environment interactions, and the resulting ultra-long trajectory (up to millions of tokens per trajectory) exceed single-GPU memory capacity, necessitating advanced parallelism strategies to enable end-to-end policy training. To address these challenges, we employ a strict on-policy, asynchronous RL training framework built on top of verl~\citep{sheng2024hybridflow}, illustrated in Figure~\ref{fig: rl_agentloop}, with two key optimizations:
\vspace{-0.5em}
\begin{itemize}[leftmargin=*]
    \item \textbf{Asynchronous Rollout for Multi-Turn Efficiency.} We implement a custom agent loop that asynchronously dispatches requests to a group of inference servers hosting the latest policy model, thereby mitigating GPU idling during environment interactions. The agent loop further incorporates asynchronous environment interaction with session management, maintaining backup sessions for seamless failover and replacement. On the server side, we employ load balancing and prefill caching to accelerate generation efficiency in multi-turn settings.
    \item \textbf{Hybrid Parallelism for Ultra-Long Sequences.} To support end-to-end training of trajectories with millions of tokens, we leverage Megatron's hybrid multi-dimensional parallelism (TP+PP+CP) to shard each long rollout trajectory across GPUs along the tensor, pipeline, and context dimensions, enabling scalable training while keeping per-GPU memory bounded. Additionally, we resize images to half their original resolution, which significantly improves training efficiency without compromising model performance.
\end{itemize}
\vspace{-0.8em}
\paragraph{Task and Verifier Design.} Effective RL training requires a well-structured task distribution that balances exploration and exploitation. We manually curate a diverse set of over 35 applications spanning simple single-app operations to complex multi-app workflows. Tasks are dynamically stratified into four difficulty levels based on the current policy's pass@K success rate (SR): \textit{frontier tasks} (0--25\% SR) push model capability boundaries, \textit{exploration tasks} (25--50\% SR) drive skill development, \textit{near-mastery tasks} (50--75\% SR) approach proficiency, and \textit{exploitation tasks} (75--100\% SR) reinforce learned behaviors.
\vspace{-0.2em}

Building on this stratification, we implement an automatic curriculum that progressively adjusts task sampling throughout training. Early stages emphasize simpler tasks to establish foundational skills, while the distribution gradually shifts toward challenging tasks as success rates improve. This adaptive strategy prevents training collapse from excessive difficulty while ensuring continuous learning signals, effectively addressing the exploration-exploitation tradeoff.

\vspace{-0.2em}
To enable scalable evaluation, we develop a hybrid verification approach tailored to task characteristics. Deterministic tasks with clear success criteria use rule-based verifiers with root-level AVD access for precise state verification. For complex tasks where rule-based verification is labor-intensive, we employ an MLLM-as-a-Judge framework to evaluate execution trajectories against task objectives. This hybrid approach achieves 83\% agreement with human annotations, enabling reliable large-scale verification without manual bottlenecks.
\vspace{-0.5em}
\paragraph{Training Algorithm.} we adopt a tailored GRPO \cite{} to sample a group of outputs $\{o_i\}_{i=1}^G$ for each task $q$, and optimizes the policy via the following objective:
\vspace{-0.5em}
{\small
\begin{equation*}
\mathcal{J}_{\text{GRPO}}(\theta) = \mathbb{E}_{(q)\sim\mathcal{D},\{o_i\}_{i=1}^G\sim\pi_{\theta_{\text{old}}}(\cdot|q)}
\left[
\frac{1}{\sum_{c=1}^G |o_c|} \sum_{i=1}^G \sum_{t=1}^{|o_i|} \min\left(r_{i,t}(\theta)\hat{A}_{i,t}, \text{clip}\left(r_{i,t}(\theta), 1-\varepsilon_{\text{low}}, 1+\varepsilon_{\text{high}}\right)\hat{A}_{i,t}\right)
\right]
\end{equation*}
}
where $\quad r_{i,t}(\theta) = \frac{\pi_{\theta}(o_{i,t} \mid q, o_{i,<t})}{\pi_{\theta_{\text{old}}}(o_{i,t} \mid q, o_{i,<t})} \quad $ is the importance sampling ratio, and $\hat{A}_{i,t} = \frac{R_i - \text{mean}(\{R_i\}_{i=1}^G)}{\text{std}(\{R_i\}_{i=1}^G)}$ is the normalized advantage. We find a group size of 16 strikes a good balance between effectiveness and efficiency.

We additionally incorporate the following features to encourage exploration and improved stability:
\begin{itemize}[leftmargin=*]
\item{\textbf{Reward Design.}} The reward signal comprises two components: a task completion reward and an action-level repetition penalty. Task completion is measured as a binary indicator of successful execution, determined by either a rule-based verifier or the MLLM-as-a-Judge framework described above. To discourage unproductive looping, we penalize recurring action sequences (from single repeated actions to cyclic patterns of 3-5 actions). Actions with identical types but different parameters are not penalized, enabling flexible execution while preventing non-progressive behavioral loops.

\item{\textbf{Clip Higher.}} Following DAPO \citep{yu2025dapo}, we employ the token-level loss with no KL divergence and an asymmetric clipping strategy with a larger upper bound to encourage exploration. Specifically, we set $\varepsilon_{\text{low}}$ to 0.2 and $\varepsilon_{\text{high}}$ to 0.3.

\item{\textbf{Experience Replay.}} We maintain a replay buffer of successful trajectories collected during training. When a rollout group contains no successful completions, we augment it with randomly sampled trajectories from the buffer. The buffer is continuously updated with newly successful experiences, retaining only the most recent eight trajectories per task to maintain near on-policy learning. This mechanism ensures  continuous learning signals even during challenging exploration phases, stabilizing training and accelerating convergence.

\end{itemize}

\subsection{Device-Cloud Collaboration}
\begin{figure}
    \centering  \includegraphics[width=0.95\linewidth]{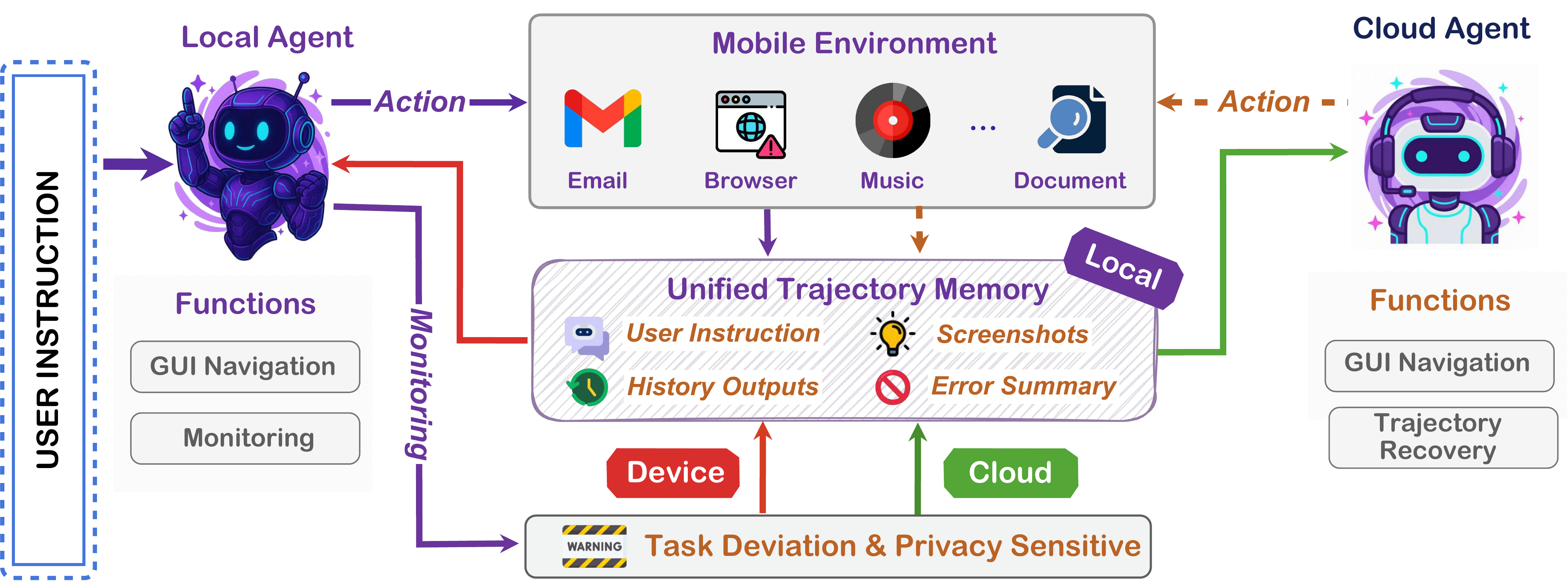}
    \vspace{-0.3em}
    \caption{Overview of device-cloud collaboration architecture. The system adaptively routes computation between device and cloud models based on task context and data sensitivity. }
    \label{fig: device_cloud_structure}
\end{figure}
\label{sec:device-cloud collaboration}
Building on the above training process, we can obtain high-capacity agents for cloud serving and lightweight yet capable on-device agents. However, neither mode alone fully meets the requirements of real-world deployment. On-device solutions are constrained by model size and thus exhibit limited GUI agent capability. Cloud deployments suffer from high latency, privacy risks, and network dependence. To address these limitations, we introduce a native device-cloud collaborative architecture that adaptively routes computation between device and cloud based on task context and data sensitivity. 
\subsubsection{System Architecture}
The overall structure of our device-cloud collaboration system is illustrated in Figure \ref{fig: device_cloud_structure}. The system consists of a Local GUI Agent, which exhibit the ability of both GUI agent and trajectory monitor, a Cloud GUI Agent, and a Local Unified Trajectory Memory that maintains consistent information exchange between local and cloud agents. The specific function of these modules are demonstrated below:
\vspace{-0.5em}
\begin{itemize}[leftmargin=*]
    \item \textbf{Local Agent.} The Local Agent runs on device and functions as both a GUI agent and a monitor. As a GUI agent, it perceives the current screen, and generates actions for each step of the task. As a monitor, it evaluates whether the trajectory so far remains aligned with the user instruction. The monitor checks indicators such as action execution failure, repeated actions without progress, incorrect inputs, or general task deviations. If the trajectory deviates from user instruction and the current context does not contain privacy-sensitive data, the monitor triggers a switch to the Cloud Agent. Additionally, the monitor generates an \textit{error summary} each time a deviation is detected, which we find highly effective for trajectory recovery. 
    \item \textbf{Cloud Agent.} The Cloud Agent is called only when the monitor detects trajectory deviation. In addition to the standard GUI agent inputs, it receives an error summary from the monitor that explains why the switch was triggered. Given the trajectory history and error summary, the Cloud Agent executes subsequent steps, leveraging its higher capacity to complete the task.
    \item \textbf{Local Unified Trajectory Memory.} On device, we maintain a unified history that records the task instruction, historical screenshots, and the model’s past outputs, including thoughts and actions. During action execution, the memory module projects the unified history into the action spaces expected by the device and cloud models, enabling either model can resume from any state without ambiguity. 
    \item \textbf{Execution loop.} The user provides a task instruction to the Local Agent. At each step, the Local Agent observes the current screenshot, decides an action, and executes it. The environmental observations and model outputs are then written to the Local Unified Trajectory Memory. Every few steps, the local agent assesses alignment between user instruction and the trajectory so far. If alignment is met, the loop continues on device. If deviation is detected and no sensitive data is involved, the system calls the Cloud Agent for task completion.
\end{itemize}
\subsubsection{Local Agent Training}
We train a single on-device model that unifies two roles: an agent for GUI navigation, and a trajectory monitor for alignment assessment. Compared with prior work \citep{light_agent}, our Local Agent introduces two key innovations required for practical device-side deployment:
\vspace{-0.5em}
\begin{itemize}[leftmargin=*]
    \item \textbf{Integrated monitoring capability.} In practice, the monitor must handle varied and complex cases, and prompt engineering alone is unlikely to deliver reliable monitoring. The on-device model is explicitly trained to judge whether the trajectory so far remains aligned with the user instruction across diverse apps, layouts, and tasks.
    \item \textbf{Error feedback generation.} When deviation is detected, our model further generates a concise error summary to guide trajectory recovery. This signal is crucial for the Cloud Agent to complete the task correctly, since handoffs occur only after the trajectory has already deviated from the user instruction.
\end{itemize}
\vspace{-1em}
\paragraph{Training procedure.} We train the Local Agent jointly on two data sources: (i) standard GUI agent data covering perception, grounding, and navigation, and (ii) monitor data that include alignment reasoning, alignment decision, and error summaries. This multi-task training recipe teaches the on-device model to execute and monitor simultaneously, without requiring separate models or fragile prompt engineering.
\vspace{-0.5em}

\subsection{MobileWorld Benchmark}
Existing benchmarks for mobile GUI agents often fail to capture the complexity of real-world mobile usage. Many evaluations rely on simple applications or limited app categories, and they typically assume idealized interaction models where user instructions are perfectly clear and agents operate solely through GUI manipulation~\citep{rawles2024androidworld,xu2025androidlab}. This gap between benchmark performance and real-world utility has become increasingly apparent. 
To evaluate \modelname{}'s practical capabilities, we adopt our \mobileworld{} \citep{mobile_world} benchmark, a comprehensive benchmark designed to bridge this evaluation gap. \mobileworld{} features over 200 realistic tasks spanning 15+ open-source applications across critical domains including e-commerce (Mall4Uni, mirroring Temu/Amazon), enterprise communication (Mattermost, mirroring Microsoft Teams/Slack), social media (Mastodon, mirroring X/Twitter), and daily productivity tools. 

\begin{table*}[!t]
    \centering
    \small
    \setlength{\tabcolsep}{2.5pt} 
    \caption{Performance comparison on the \textbf{ScreenSpot-Pro} benchmark. We use `$^*$' to denote the results evaluated by us. The best results are highlighted in \textbf{bold}, and the second-best results are \underline{underlined}.}
    \label{tab:main_results_screenspot_pro}

    \begin{tabularx}{\textwidth}{
        l
        *{12}{>{\centering\arraybackslash}X}
        c}
        \toprule
        \multirow{2}{*}{\textbf{Model}} &
        \multicolumn{2}{c}{\textbf{CAD}} &
        \multicolumn{2}{c}{\textbf{Dev.}} &
        \multicolumn{2}{c}{\textbf{Creative}} &
        \multicolumn{2}{c}{\textbf{Scientific}} &
        \multicolumn{2}{c}{\textbf{Office}} &
        \multicolumn{2}{c}{\textbf{OS}} &
        \multirow{2}{*}{\textbf{Avg.}} \\
        \cmidrule(lr){2-3}
        \cmidrule(lr){4-5}
        \cmidrule(lr){6-7}
        \cmidrule(lr){8-9}
        \cmidrule(lr){10-11}
        \cmidrule(lr){12-13}
        & Text & Icon & Text & Icon & Text & Icon & Text & Icon & Text & Icon & Text & Icon & \\
        \midrule
        \rowcolor{gray!15}
        \multicolumn{14}{l}{\textit{Proprietary Models}} \\
        GPT-4o~\citep{gpt4o}
            & 2.0  & 0.0  & 1.3  & 0.0  & 1.0  & 0.0  & 2.1  & 0.0
            & 1.1  & 0.0  & 0.0  & 0.0  & \cellcolor{light_purple}0.8 \\

        Claude C.~\citep{claude_com}
            & 14.5 & 3.7  & 22.0 & 3.9  & 25.9 & 3.4  & 33.9 & 15.8
            & 30.1 & 16.3 & 11.0 & 4.5  & \cellcolor{light_purple}17.1 \\

        Gemini-3-Pro~\citep{gemini3pro} 
            & -    & -    & -    & -    & -    & -    & -    & -
            & -    & -    & -    & -    & \cellcolor{light_purple}72.7 \\
        Seed1.8~\citep{seed18}& -    & -    & -    & -    & -    & -    & -    & -
            & -    & -    & -    & -    & \cellcolor{light_purple}\underline{73.1} \\
        \midrule
        \rowcolor{gray!15}
        \multicolumn{14}{l}{\textit{Open-Source Models}} \\
        Qwen3-VL-2B$^*$~\citep{Qwen3-VL} &31.0 &15.6 &55.2 &11.7 &59.1 &16.1 &64.6 &22.7 &72.3 &34.0 &59.8 &23.6 &\cellcolor{light_purple}41.9 \\
        InfiGUI-3B \citep{infiguig1} & 50.8 & 25.0 & 64.9 & 20.0 & 51.5 & 16.8 & 68.8 & 32.7 & 70.6 & 32.1 & 49.5 & 19.7 & \cellcolor{light_purple}45.2 \\ 
        Ferret-UI Lite \citep{Ferret-ui-lite}& -    & -    & -    & -    & -    & -    & -    & -
            & -    & -    & -    & -    & \cellcolor{light_purple}53.3\\

        UI-TARS-7B~\citep{ui-tars}
            & 20.8 & 9.4  & 58.4 & 12.4 & 50.0 & 9.1  & 63.9 & 31.8
            & 63.3 & 20.8 & 30.8 & 16.9 & \cellcolor{light_purple}35.7 \\
        Phi-Ground~\citep{phi_ground}
            & 26.9 & 17.2 & 70.8 & 16.7 & 56.6 & 13.3 & 58.0 & 29.1 & 76.4 & 44.0 & 55.1 & 25.8 & \cellcolor{light_purple}43.2\\
        GUI-Actor-7B~\citep{gui_actor}
            & 47.7 & 9.4  & 59.1 & 15.9 & 59.6 & 16.1 & 70.1 & 25.5
            & 69.5 & 41.5 & 55.1 & 19.1 & \cellcolor{light_purple}44.6 \\
        SE-GUI-7B~\citep{SE_GUI}
            & 51.3 & 14.1 & 68.2 & 19.3 & 57.6 & 9.1  & 75.0 & 28.2
            & 78.5 & 43.4 & 49.5 & 25.8
            & \cellcolor{light_purple}47.2 \\
        GUI-G$^2$-7B~\citep{GUI_G2}
            & 55.8 & 12.5 & 68.8 & 17.2 & 57.1 & 15.4 & 77.1 & 24.5
            & 74.0 & 32.7 & 57.9 & 21.3 & \cellcolor{light_purple}47.5 \\
        Qwen3-VL-8B$^*$~\citep{Qwen3-VL} & 46.7 &10.9 &79.2 &23.4 &68.2 &14.0 &73.6 &30.0 &76.3 &30.2 &65.4 &21.3 &\cellcolor{light_purple}49.9\\
        OpenCUA-7B~\citep{opencua}
            & -    & -    & -    & -    & -    & -    & -    & -
            & -    & -    & -    & -    & \cellcolor{light_purple}50.0 \\
        GTA1-7B~\citep{GTA1}
            & 53.3 & 17.2 & 66.9 & 20.7 & 62.6 & 18.9
            & 76.4 & 31.8 & 82.5 & 50.9 & 48.6 & 25.9
            & \cellcolor{light_purple}50.1 \\

        UI-Venus-7B~\citep{ui_venus}
            & 60.4 & 21.9 & 74.7 & 24.1 & 63.1 & 14.7
            & 76.4 & 31.8 & 75.7 & 41.5 & 49.5 & 22.5
            & \cellcolor{light_purple}50.8 \\
        InfiGUI-G1-7B~\citep{infiguig1}
            & 57.4 & 23.4 & 74.7 & 24.1 & 64.6 & 18.2 & 80.6 & 31.8
            & 75.7 & 39.6 & 57.0 & 29.2 & \cellcolor{light_purple}51.9 \\
        GUI-Owl-7B~\citep{mobileagentv3}
            & 64.5 & 21.9 & 76.6 & 31.0 & 59.6 & 27.3 & 79.1 & 37.3 & 77.4 & 39.6 & 59.8 & 33.7 & \cellcolor{light_purple}54.9 \\
        \midrule
        Qwen3-VL-32B$^*$~\citep{Qwen3-VL} & 60.4 &28.1 &69.5 &22.1 &75.8 &25.2 &84.7 &25.5 &85.9 &43.4 &62.6 &15.7 &\cellcolor{light_purple}54.9\\

        OpenCUA-32B~\citep{opencua}
            & -    & -    & -    & -    & -    & -    & -    & -
            & -    & -    & -    & -    & \cellcolor{light_purple}55.3 \\

        GUI-Owl-32B~\citep{mobileagentv3}
            & 62.4 & 28.1 & \underline{84.4} & 39.3 & 65.2 & 18.2 & 82.6 & 39.1 & 81.4 & 39.6 & 70.1 & 36.0 & \cellcolor{light_purple}58.0 \\
        GTA1-32B~\citep{GTA1}
            & 43.7 & 23.4 & 82.5 & 28.3
            & 69.2 & 14.7 & 79.9 & 31.8
            & 80.8 & 43.4 & 70.1 & 32.6
            & \cellcolor{light_purple}63.6 \\
        UGround-v1-72B~\citep{uground}
            & 16.8 & 4.7  & 55.8 & 4.8  & 54.0 & 10.5 & 70.8 & 22.7
            & 61.0 & 18.9 & 40.2 & 7.9  & \cellcolor{light_purple}34.5 \\
        UI-Tars-72B~\citep{ui-tars}
            & 18.8 & 12.5 & 63.0 & 17.2 & 57.0 & 15.4 & 64.6 & 20.9
            & 63.3 & 26.4 & 42.1 & 15.7 & \cellcolor{light_purple}38.1 \\

        UI-Venus-72B~\citep{ui-venus}
            & 66.5 & 29.7 & 84.4 & 33.1 & 73.2 & 30.8 & 84.7 & 42.7 & 83.1 & 60.4 & 75.7 & 36.0  & \cellcolor{light_purple}61.9\\

        \midrule
        \rowcolor{gray!15}
        \multicolumn{14}{l}{\textit{Ours}} \\
        \textbf{\modelname-2B}&61.4 &23.4 &76.6 &32.4 &69.2 &21.7 &81.2 &34.5 &85.9 &39.6 &68.2 &41.6 & \cellcolor[HTML]{cfcdfd}57.4\\
        \rowcolor{gray!10}
        \textit{+ Zoom-In} & 69.5 &34.4 &75.3 &42.8 &74.7 &30.1 &84.0 &42.7 &85.3 &56.6 &69.2 &47.2 & \cellcolor[HTML]{cfcdfd}62.8\\
        
        \textbf{\modelname-8B}&72.6 &35.9 &83.8 &52.4 &76.3 &33.6 &79.9 &37.3 & \underline{88.7} &60.4 &76.6 & \underline{49.4} & \cellcolor[HTML]{cfcdfd}65.8\\
        \rowcolor{gray!10}
        \textit{+ Zoom-In} & \textbf{80.7} &43.8 &78.6 & \textbf{58.6} &78.8 & \textbf{46.9} & \underline{86.1} & \underline{49.1} &88.1 & \textbf{81.1} &76.6 & \textbf{51.7} & \cellcolor[HTML]{cfcdfd}70.9\\

        \textbf{\modelname-32B}&70.1 & \underline{45.3} & \textbf{86.4} &40.7 & \underline{82.8} &37.8 & \textbf{91.7} &46.4 & \textbf{90.4} &71.7 & \underline{78.5} &34.8 & \cellcolor[HTML]{cfcdfd}67.9\\
        \rowcolor{gray!10}
        \textit{+ Zoom-In}& \underline{79.2} &\textbf{53.1} & \underline{84.4} & \underline{57.9} & \textbf{87.9} & \underline{46.2} & \textbf{91.7} & \textbf{54.5} &88.1 & \underline{79.2} & \textbf{80.4} &47.2 & \cellcolor[HTML]{cfcdfd}\textbf{73.5} \\
        \bottomrule
    \end{tabularx}
    \vspace{-0.8em}
\end{table*}

In addition, \mobileworld{} extends beyond standard GUI manipulation to evaluate two essential real-world capabilities, which directly align with \modelname{}'s design of practical deployment.
\vspace{-0.5em}
\begin{itemize}[leftmargin=*]
    \item \textbf{Agent-User Interaction}, where agents must detect ambiguous user requests and proactively seek clarification rather than making incorrect assumptions.
    \item \textbf{MCP Tool Integration}, where agents must intelligently decide between manual GUI navigation and API-based operations via MCP~\citep{anthropic_mcp_intro} to optimize efficiency.
\end{itemize}

\section{Experiments}
\vspace{-0.5em}
We present comprehensive experiments in this section, including extensive benchmark evaluations and detailed ablations and analyses of the main components of \modelname{}. Specifically, we introduce the experimental setup (Sec. \ref{exp_setup}); report the main benchmark results across GUI grounding, GUI navigation, and real-world-oriented evaluation (Sec. \ref{main_result}); present case studies of MCP augmentation and agent–user interaction (Sec. \ref{mcp_call_user}); demonstrate experiments and case studies on the native device–cloud collaboration system (Sec. \ref{device_cloud_exp_sec}); discuss online RL ablations (Sec. \ref{online_rl_exp_sec}); and, finally, provide grounding analysis (Sec. \ref{sec:grounding_analysis}).
\subsection{Experimental Setup}
\label{exp_setup}
\vspace{-0.5em}
\paragraph{Implementation details}
\modelname{} uses Qwen3-VL \citep{Qwen3-VL} as the backbone across multiple model sizes (2B, 8B, 32B, and a 235B-A22B) to meet realistic deployment needs. Training proceeds in four stages: (i) SFT on perception and grounding data, (ii) SFT on mobile-use navigation data with a small portion of grounding data, (iii) RL for grounding to develop robust UI grounding capability, and (iv) online RL for mobile-use navigation in dynamic environments to enhance robustness and generalization. To further enhance the large cloud model variants, we augment the training of the 32B and 235B‑A22B models with more real-world trajectories.
\begin{table}[t!]
\centering
\footnotesize
\small
\setlength{\tabcolsep}{16pt}
\caption{Performance comparison on \textbf{UI-Vision} grounding dataset. The best results are highlighted in \textbf{bold}, and the second-best results are \underline{underlined}.}

\begin{tabular}{lcccc}
\toprule
\textbf{Models} & \textbf{Basic} & \textbf{Functional} & \textbf{Spatial}& \textbf{Avg}\\ 
\midrule
\rowcolor{gray!15}
\multicolumn{5}{l}{\textit{Proprietary Models}} \\
GPT-4o~\citep{gpt4o}& 1.6 & 1.5 &1.0&\cellcolor{light_purple}1.4\\ 
Claude-3.7-Sonnet~\citep{claude37}& 9.5 & 7.7&7.6&\cellcolor{light_purple}8.3 \\
\midrule
\rowcolor{gray!15}
\multicolumn{5}{l}{\textit{Open-source Models}} \\
Qwen3-VL-2B$^*$~\citep{Qwen3-VL} &0.0 &19.2 &0.1 &\cellcolor{light_purple}6.2 \\
InfiGUI-G1-3B~\citep{infiguig1} & 31.2 & 28.0 & 8.2 & \cellcolor{light_purple}22.0\\
Qwen2.5-VL-7B~\citep{qwen25vl}&1.2& 0.8&0.5&\cellcolor{light_purple}0.9\\ 
SeeClick~\citep{seeclick}&9.4 & 4.7&2.1&\cellcolor{light_purple}5.4 \\ 
UGround-V1-7B~\citep{uground}&15.4& 17.1& 6.3 &\cellcolor{light_purple}12.9\\ 
OS-Atlas-7B~\citep{osaltas_and_screenspot_v2}&12.2 &11.2 &3.7 &\cellcolor{light_purple}9.0\\
Qwen3-VL-8B$^*$~\citep{Qwen3-VL} &25.0 &27.9 &1.2 &\cellcolor{light_purple}17.5 \\
UI-TARS-7B~\citep{uitars}& 20.1 &24.3 &8.4 &\cellcolor{light_purple}17.6\\

UI-TARS-1.5-7B~\citep{ui-tars-15}& 28.8 &27.5& 10.7 &\cellcolor{light_purple}22.3 \\
InfiGUI-G1-7B~\citep{infiguig1} & 36.2 & 31.9 & 11.5 & \cellcolor{light_purple}26.1 \\
UI-Venus-7B~\citep{ui_venus} &36.1&32.8 &11.9&\cellcolor{light_purple}26.5\\ 
Phi-Ground~\citep{phi_ground}&36.8 &37.1 & 7.6 &\cellcolor{light_purple}27.2\\
Qwen3-VL-32B$^*$~\citep{Qwen3-VL} & 32.8 &34.2 &14.7 &\cellcolor{light_purple}26.9\\
UI-TARS-72B~\citep{ui-tars}& 31.4 &30.5 & 14.7 & \cellcolor{light_purple}25.5\\
UI-Venus-72B~\citep{ui_venus}& 45.6 & 42.3 & 23.7&\cellcolor{light_purple}36.8\\ 
\midrule
\rowcolor{gray!15}
\multicolumn{5}{l}{\textit{Ours}} \\
\textbf{\modelname-2B} & 41.0 &41.2 &10.4 &\cellcolor[HTML]{cfcdfd}30.3\\
\rowcolor{gray!10}
\textit{+ Zoom-In} & 43.2 &43.0 &11.3 &\cellcolor[HTML]{cfcdfd}31.9\\

\textbf{\modelname-8B} & 51.7 &49.6 &22.5 & \cellcolor[HTML]{cfcdfd}40.7\\
\rowcolor{gray!10}
\textit{+ Zoom-In} & 51.6 &50.5 &26.6 & \cellcolor[HTML]{cfcdfd}42.4\\

\textbf{\modelname-32B} & \textbf{59.1} &\textbf{57.1} &\underline{26.9} &\cellcolor[HTML]{cfcdfd}\underline{47.1}\\
\rowcolor{gray!10}
\textit{+ Zoom-In} & \underline{58.7} &\underline{56.8} &\textbf{33.6} &\cellcolor[HTML]{cfcdfd}\textbf{49.2}\\
\bottomrule
\end{tabular}
\label{tab:ui-vison}
    \vspace{-0.8em}
\end{table}
\vspace{-0.5em}
\paragraph{Benchmarks}
We evaluate \modelname{} across extensive benchmarks spanning three categories: grounding, mobile-use, and real-world–oriented evaluation.
\begin{itemize} [leftmargin=*]
    \item \textbf{Grounding benchmarks.} We evaluate grounding with five complementary benchmarks: ScreenSpot-Pro \citep{li2025screenspotpro} for high-resolution, fine-grained professional layouts, UI-Vision \citep{ui_vision} for diverse applications and reasoning types (e.g., spatial and functional), MMBench-GUI L2 \citep{mmbenchgui} for hierarchical-instruction following and compositional reasoning, OSWorld-G and OSWorld-G Refine \citep{osworld_g} for comprehensive skills such as layout understanding, widget matching, and fine-grained manipulation, and ScreenSpot-V2 \citep{osaltas_and_screenspot_v2} to broaden coverage across different operating systems. 
    \item \textbf{Mobile-use benchmarks.} We report offline and online results. Offline evaluation includes Android Control \citep{android_control}, which evaluates planning and action execution capabilities in the mobile setting, and GUI Odyssey \citep{gui_odyssey}, which covers cross-app navigation tasks. For online evaluation, AndroidWorld~\citep{android_world} provides 116 tasks across 20 Android apps in a live Android emulator, and requires continuous interaction with dynamic mobile environment.
    \item \textbf{Real-world–oriented evaluation.} We introduce MobileWorld \citep{mobile_world}, a more challenging, dynamic benchmark that closely mirrors production usage. It includes tasks requiring agent–user interaction and MCP tool use, enabling rigorous evaluation of these two capabilities that are critical in real-world settings.
\end{itemize}

\begin{table*}[t!]
    \centering
    \small
    \setlength{\tabcolsep}{1.6pt}
    \caption{Performance comparison on the \textbf{MMBench-GUI L2} benchmark. The best results are highlighted in \textbf{bold}, and the second-best results are \underline{underlined}.}
    \label{tab:main_results_mmbench}

    \begin{tabularx}{\textwidth}{
        l
        *{12}{>{\centering\arraybackslash}X}
        c}
        \toprule
        \multirow{2}{*}{\textbf{Model}} &
        \multicolumn{2}{c}{\textbf{Windows}} &
        \multicolumn{2}{c}{\textbf{MacOS}} &
        \multicolumn{2}{c}{\textbf{Linux}} &
        \multicolumn{2}{c}{\textbf{iOS}} &
        \multicolumn{2}{c}{\textbf{Android}} &
        \multicolumn{2}{c}{\textbf{Web}} &
        \multirow{2}{*}{\textbf{Avg.}} \\
        \cmidrule(lr){2-3}
        \cmidrule(lr){4-5}
        \cmidrule(lr){6-7}
        \cmidrule(lr){8-9}
        \cmidrule(lr){10-11}
        \cmidrule(lr){12-13}
        & Bas. & Adv. & Bas. & Adv. & Bas. & Adv. & Bas. & Adv. & Bas. & Adv. & Bas. & Adv. & \\
        \midrule
        \rowcolor{gray!15}
        \multicolumn{14}{l}{\textit{Proprietary Models}} \\
        GPT-4o~\citep{gpt4o}
            & 1.5 & 1.1 & 8.7 & 4.3 & 1.1 & 1.0 & 5.1 & 3.3 & 2.5 & 1.4 & 3.2 & 2.9
            & \cellcolor{light_purple}2.9 \\
        Claude-3.7~\citep{claude37}
            & 1.5 & 0.7 & 12.5 & 7.5 & 1.1 & 0.0 & 13.7 & 10.6 & 1.4 & 1.4 & 3.2 & 2.3
            & \cellcolor{light_purple}4.7 \\
        Qwen-Max-VL~\citep{qwen2}
            & 43.9 & 36.8 & 58.8 & 56.1 & 53.9 & 30.1 & 77.4 & 59.1 & 79.5 & 70.1 & 74.8 & 58.8
            & \cellcolor{light_purple}58.0 \\
        \midrule
        \rowcolor{gray!15}
        \multicolumn{14}{l}{\textit{Open-Source Models}} \\
        Qwen3-VL-2B$^*$~\citep{Qwen3-VL}
            & 81.9 &0.0 &80.3 &46.2 &67.5 &0.0 &90.8 &0.0 &91.0 &0.0 &88.1 &0.0 &\cellcolor{light_purple}46.5\\
        OS-Atlas-7B~\citep{osaltas_and_screenspot_v2}
            & 36.9 & 18.8 & 44.4 & 21.7 & 31.4 & 13.3 & 74.8 & 48.8 & 69.6 & 46.8 & 61.3 & 35.4
            & \cellcolor{light_purple}41.4 \\
        Aguvis-7B~\citep{aguvis}
            & 37.3 & 21.7 & 48.1 & 33.3 & 33.5 & 25.0 & 67.5 & 65.2 & 61.0 & 51.0 & 61.6 & 45.5
            & \cellcolor{light_purple}45.7 \\
        UI-TARS-1.5-7B~\citep{ui-tars-15}
            & 68.3 & 39.0 & 69.0 & 44.5 & 64.4 & 37.8 & 88.5 & 69.4 & 90.5 & 69.3 & 81.0 & 56.5
            & \cellcolor{light_purple}64.3 \\
        UGround-V1-7B~\citep{uground}
            & 66.8 & 39.0 & 71.3 & 48.6 & 56.5 & 31.1 & 92.7 & 70.9 & 93.5 & 71.0 & 88.7 & 64.6
            & \cellcolor{light_purple}65.7 \\
        GUI-Actor-7B$^*$~\citep{gui_actor}
            & 80.8 & 55.1 & 81.4 & 60.4 & 64.9 & 41.8 & 94.3 & 82.7 & 93.5 & 79.7 & 89.7 & 72.1
            & \cellcolor{light_purple}76.5 \\
        SE-GUI-7B$^*$~\citep{SE_GUI}
            & 77.5 & 57.7 & 77.1 & 60.7 & 68.6 & 44.9 & 95.5 & 80.0 & 95.5 & 83.7 & 89.7 & 68.8
            & \cellcolor{light_purple}76.6 \\
        Qwen3-VL-8B$^*$~\citep{Qwen3-VL} 
            & 88.6 &61.8 &85.5 &69.1 &74.9 &53.1 & 95.2 &82.4 &95.5 &84.5 & \textbf{96.8} &72.1 &\cellcolor{light_purple}81.3\\
        GTA1-7B$^*$~\citep{GTA1}
            & 76.8 & 57.4 & 80.3 & 63.9 & 68.6 & 53.6 & 93.9 & 83.3 & 96.3 & 84.5 & 90.3 & 74.7
            & \cellcolor{light_purple}78.5 \\
        GUI-G$^2$-7B$^*$~\citep{GUI_G2}
            & 79.7 & 55.1 & 79.7 & 64.7 & 69.6 & 50.0 & 95.2 & 82.7 & 96.6 & 85.4 & 91.9 & 75.6
            & \cellcolor{light_purple}78.8 \\
        GUI-Owl-7B ~\citep{mobileagentv3} 
            & 86.4 & 61.8 & 81.7 & 64.5 & 74.4 & 61.7
            & 94.9 & 83.0 & 95.8 & 83.7 & 93.2 & 72.7
            & \cellcolor{light_purple}80.5 \\
        InfiGUI-G1-7B~\citep{infiguig1}
            & 82.7 & 61.8 & 83.8 & 63.9 & 72.3 & 52.0 & 94.9 & 89.4 & 95.2 & 85.6 & 93.5 & 76.3
            & \cellcolor{light_purple}80.8 \\

        \midrule


        GUI-Owl-32B ~\citep{mobileagentv3} 
            & 85.6 & 65.1 & 84.9& 67.1 & 77.0 & 63.3
            & 95.2 & 85.5 & 96.1 & 87.0 & 95.5 & 80.8
            & \cellcolor{light_purple}83.0 \\
        GTA1-32B$^*$~\citep{GTA1}
            & 82.3 & 66.9 & 89.0 & 74.0 & 73.3 & 52.0
            & \underline{96.2} & 88.2 & 95.8 & 88.5 & 95.2 & 79.9
            & \cellcolor{light_purple}83.4 \\
        Qwen3-VL-32B$^*$~\citep{Qwen3-VL} 
            & \textbf{93.4} &71.3 & \textbf{92.8} &74.3 &78.0 &56.1 &95.5 &88.8 &97.2 &88.5 &92.6 &78.6 &\cellcolor{light_purple}85.3\\

        UI-TARS-DPO-72B~\citep{ui-tars}
            & 78.6 & 51.8 & 80.3 & 62.7 & 68.6 & 51.5 & 90.8 & 81.2 & 93.0 & 80.0 & 88.1 & 68.5
            & \cellcolor{light_purple}74.3 \\
        InternVL3-78B~\citep{internvl3}
            & 70.1 & 42.6 & 75.7 & 52.3 & 59.2 & 41.3 & 93.6 & 80.6 & 92.7 & 78.6 & 90.7 & 65.9
            & \cellcolor{light_purple}72.2 \\

        \midrule
        \rowcolor{gray!15}
        \multicolumn{14}{l}{\textit{Ours}} \\
        \textbf{\modelname-2B} &84.9 &64.0 &89.3 &72.5 &75.4 &60.2 &95.2 &85.2 &96.3 &84.2 &92.9 &76.0 & \cellcolor[HTML]{cfcdfd}82.6 \\
        
        \textbf{\modelname-8B}& 92.3 & \underline{74.3} & \underline{90.7} & \underline{86.4} & \underline{81.2} & \underline{67.3} & \textbf{97.1} & \underline{90.0} & \underline{97.5} & \underline{92.7} & 95.8 & \underline{86.0} & \cellcolor[HTML]{cfcdfd}\underline{88.8}\\

        \textbf{\modelname-32B}& \underline{93.0} &\textbf{78.7} &\textbf{92.8} &\textbf{87.6} &\textbf{86.9} &\textbf{77.6} &\textbf{97.1} &\textbf{92.4} &\textbf{98.0} &\textbf{93.2} &\underline{96.1} &\textbf{92.5} & \cellcolor[HTML]{cfcdfd}\textbf{91.3}\\

        \bottomrule
    \end{tabularx}
    \vspace{-0.8em}
\end{table*}

\subsection{Main Results}
\label{main_result}
\subsubsection{Grounding Capability}
\begin{figure}
    \centering
    \includegraphics[width=0.98\linewidth]{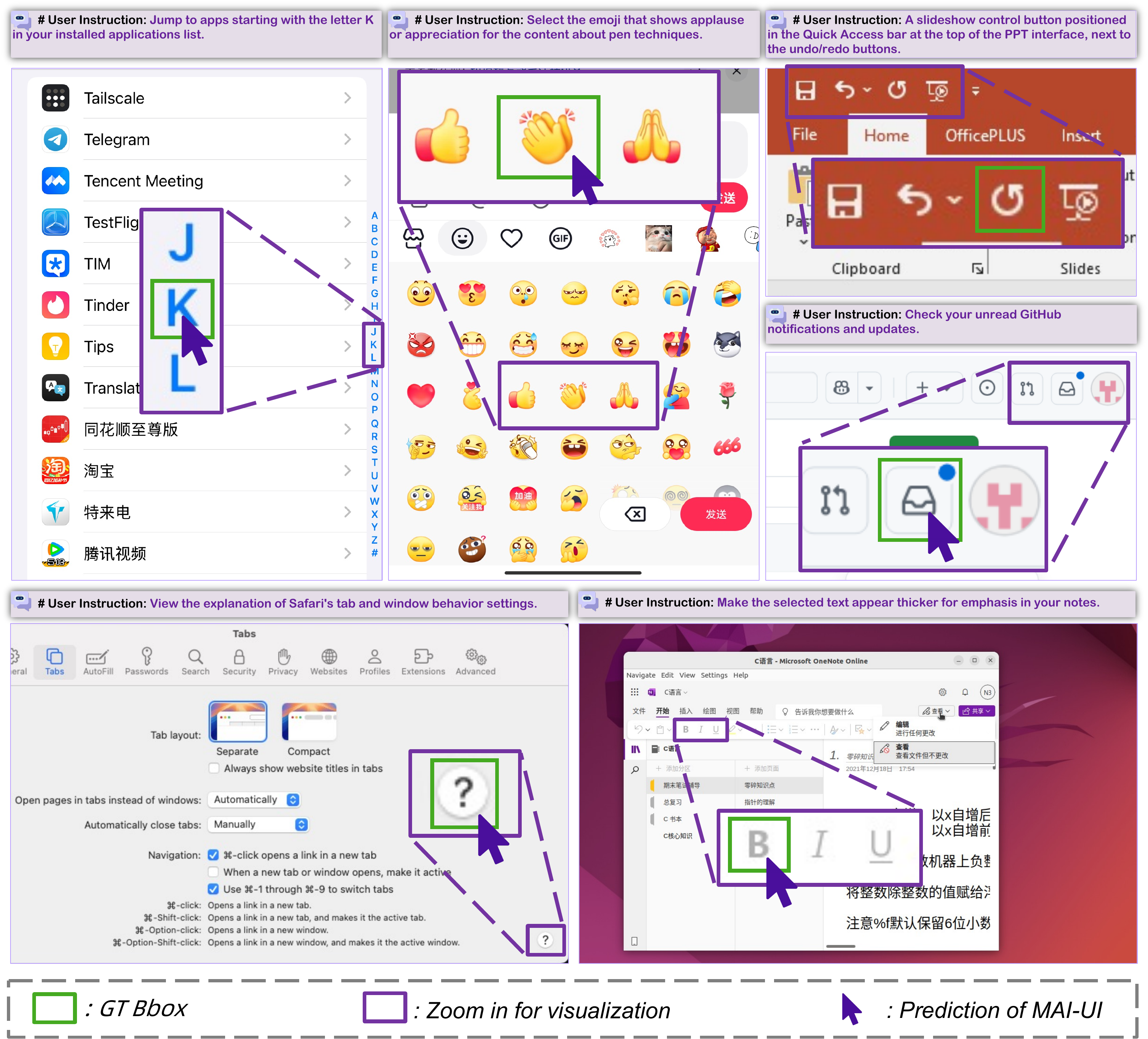}
    \caption{Grounding case studies across different operating systems.}
    \label{fig:Grounding_Cases}
\end{figure}

{\paragraph{Overall Grounding Performance. } We evaluate \modelname{} on five comprehensive GUI grounding benchmarks. Across all scales, our 2B, 8B, and 32B variants consistently outperform models of comparable size and establish new state-of-the-art results. On ScreenSpot-Pro~\citep{li2025screenspotpro}, \modelname{}-32B attains 67.9\% accuracy, an 4.1\% absolute accuracy gain over the strongest baseline GTA1-32B. With the adaptive zoom-in strategy, performance further increases to {73.5\%}, surpassing Gemini 3 Pro~\citep{gemini3pro} and Seed1.8~\citep{seed18}. On OSWorld-G~\citep{osworld_g}, our models show consistent improvements over the best comparable baselines. Specifically, \modelname{}-32B with zoom-in achieves 70.9\% on OSWorld-G and 75.0\% on OSWorld-G Refine. On UI-Vision~\citep{ui_vision}, \modelname{}-32B achieves 47.1\% accuracy, and it increases to 49.2\% with the zoom-in strategy, making an absolute gain of {+12.4} points over the previous best UI-Venus-Ground-72B. On MMBench-GUI L2~\citep{mmbenchgui}, \modelname{} reaches {91.3\%}, surpassing the prior best GTA1 by {+7.9} points. On ScreenSpot-V2~\citep{osaltas_and_screenspot_v2}, \modelname{} sets a new SOTA at 96.5\%. It is also worth noting that \modelname{}-2B, despite its small size, demonstrates strong grounding performance. For example, it achieves {62.8\%} on ScreenSpot-Pro with the zoom-in operation. This score outperforms GUI-Owl-32B and UI-Venus-72B. We present grounding case studies across different operating systems in Figure \ref{fig:Grounding_Cases}. Detailed per-benchmark results are discussed below.
}
\begin{table}[!t]
    \caption{Performance comparison of state-of-the-art models on the \textbf{OSWorld-G}. The best results are highlighted in \textbf{bold}, and the second-best results are \underline{underlined}.}
    \centering
    \setlength{\tabcolsep}{4pt}
    \small
    \begin{tabular}{lcccccc}
        \toprule
        \textbf{Agent Model} & \makecell[c]{\textbf{Text}\\\textbf{Matching}} & \makecell[c]{\textbf{Element}\\\textbf{Recognition}} & \makecell[c]{\textbf{Layout}\\\textbf{Understanding}} & \makecell[c]{\textbf{Fine-grained}\\\textbf{Manipulation}}& \textbf{Refusal} & \cellcolor{white}\textbf{Avg} \\
        \midrule
        \rowcolor{gray!15}
        \multicolumn{7}{l}{\textit{Proprietary Models}} \\
        Operator  \citep{OpenAICUA} & 51.3 & 42.4 & 46.6 & 31.5 & 0.0 & \cellcolor{light_purple}40.6 \\
        Seed1.5-VL \citep{ui-tars-15-seed} & 73.9 & 66.7 & 69.6 & 47.0 & 18.5 &\cellcolor{light_purple}62.9 \\
        \midrule
        \rowcolor{gray!15}
        \multicolumn{7}{l}{\textit{Open-Source Models}} \\
        Jedi-3B \citep{jedi} & 67.4 & 53.0 & 53.8 & 44.3 & 7.4 & \cellcolor{light_purple}50.9 \\
        Qwen3-VL-2B$^*$~\citep{Qwen3-VL}& 61.7 &45.8 &54.2 &39.6 &- &\cellcolor{light_purple}45.9\\
        OS-Atlas-7B \citep{osaltas_and_screenspot_v2}& 44.1 & 29.4 & 35.2 & 16.8 & 7.4 & \cellcolor{light_purple}27.7 \\
        UGround-7B \citep{uground} & 51.3 & 40.3 & 43.5 & 24.8 & 0.0 & \cellcolor{light_purple}36.4 \\
        Aguvis-7B \citep{aguvis} & 55.9 & 41.2 & 43.9 & 28.2 & 0.0 & \cellcolor{light_purple}38.7 \\
        UI-TARS-7B \citep{uitars} & 60.2 & 51.8 & 54.9 & 35.6 & 0.0 & \cellcolor{light_purple}47.5 \\
        UI-TARS-1.5-7B \citep{ui-tars-15} & 36.8 & 62.7 & 62.2 & 50.8 & 0.0 & \cellcolor{light_purple}52.8\\
        Jedi-7B \citep{jedi} & 65.9 & 55.5 & 57.7 & 46.9 & 7.4 & \cellcolor{light_purple}54.1 \\
        Qwen3-VL-8B$^*$~\citep{Qwen3-VL}&69.0 &55.5 &59.7 &47.7 &- &\cellcolor{light_purple}54.8\\
        GTA1-7B~\citep{GTA1} & 42.1 & 65.7 & 62.7 & 56.1 & 0.0 & \cellcolor{light_purple}55.1 \\
        GUI-Owl-7B~\citep{mobileagentv3} & 64.8 & 63.6 & 61.3 & 41.0 & - & \cellcolor{light_purple}55.9 \\
        UI-Venus-7B~\citep{ui_venus} & 74.6 & 60.5 & 61.5 & 45.5 & - & \cellcolor{light_purple}58.8 \\
        \midrule
        OpenCUA-32B \citep{opencua} & - & - & - & - & - & \cellcolor{light_purple}59.6 \\
        GUI-Owl-32B~\citep{mobileagentv3} & 67.0 & 64.5 & 67.2 & 45.6 & - & \cellcolor{light_purple}58.0 \\
        Qwen3-VL-32B$^*$~\citep{Qwen3-VL}&72.8 &63.3 &66.4 &51.7 &- &\cellcolor{light_purple}60.6\\

        GTA1-32B~\citep{GTA1} & 63.2 & \textbf{78.4} & 73.3 & \textbf{65.2} &0.0&\cellcolor{light_purple}65.2  \\

        UI-TARS-72B \citep{ui-tars}& 69.4 & 60.6 & 62.9 & 45.6 & 0.0 &\cellcolor{light_purple}57.1 \\
        UI-Venus-72B~\citep{ui_venus} & \textbf{82.1} & 71.2 & 70.7 & \underline{64.4} & - & \cellcolor{light_purple}\underline{70.4} \\

        \midrule
        \rowcolor{gray!15}
        \multicolumn{7}{l}{\textit{Ours}} \\
        \textbf{\modelname-2B} & 62.8 &56.7 &59.3 &40.3 &- & \cellcolor[HTML]{cfcdfd}52.0\\
        \rowcolor{gray!10}
        \textit{+ Zoom-In} & 66.7 &59.4 &63.2 &44.3 &- & \cellcolor[HTML]{cfcdfd}55.9\\
        
        \textbf{\modelname-8B} & 72.0 &63.3 &66.0 &51.0 &- & \cellcolor[HTML]{cfcdfd}60.1\\
        \rowcolor{gray!10}
        \textit{+ Zoom-In} & 72.8 &67.6 &71.1 &56.4 &- & \cellcolor[HTML]{cfcdfd}64.2\\

        \textbf{\modelname-32B} & 73.6 &72.4& \underline{73.9} &57.7 &- & \cellcolor[HTML]{cfcdfd}67.6\\
        \rowcolor{gray!10}
        \textit{+ Zoom-In} & \underline{78.5} & \underline{75.2} & \textbf{78.3} &62.4 &- &\cellcolor[HTML]{cfcdfd}\textbf{70.9}\\
        \bottomrule
    \end{tabular}
    \label{tab:osworld_g_comparison}
    \vspace{-0.8em}
\end{table}
\vspace{-1em}
\paragraph{Grounding for high-resolution scenario. } \textit{ScreenSpot-Pro} evaluates grounding performance on high-resolution professional software with dense and fine-grained UI layouts. As illustrated in Table \ref{tab:main_results_screenspot_pro}, \modelname{} achieves the best grounding performance across all categories, including CAD, development, creative, scientific, office, and operation systems. Averaged over categories, the \modelname{}'s 2B, 8B, and 32B variants achieve 57.4\%, 65.8\%, and 67.9\%, surpassing the best baselines of comparable size by {+4.1}, {+10.9}, and {+4.3} points, respectively. With the zoom-in strategy, \modelname-32B further attains an average accuracy of {73.5\%}, significantly surpassing strong open-source baselines such as GTA1-32B \citep{GTA1}, UI-Venus \citep{ui-venus}, and GUI-Owl \citep{mobileagentv3}, and even outperforming Gemini 3 Pro \citep{gemini3pro} and Seed1.8~\citep{seed18}. These results highlight the effectiveness of our model on complex, high-resolution screens.
\vspace{-1em}
\paragraph{Grounding for diverse and complex instruction.} To assess grounding performance in terms of instruction diversity and complexity, we use \textit{UI-Vision} and \textit{MMBench-GUI L2} for evaluation. UI-Vision provides multi-perspective queries (e.g., spatial, functional), and MMBench-GUI L2 includes instructions of Basic (low-level, attribute/appearance) and Advanced (high-level, goal-oriented) categories. Together, these benchmarks reflect realistic usage where user instructions are heterogeneous, high-level, and often implicit. On UI‑Vision, our model sets a new state of the art: \modelname{}‑32B with zoom‑in achieves a 49.2\% average accuracy, exceeding the strongest baseline (UI‑Venus‑Ground‑72B 36.8\%) by +12.4 points. The 8B and 2B variants of \modelname{} also outperform baselines of similar size by +15.2 and +9.9 points, demonstrating superior grounding performance across diverse instruction perspectives. On MMBench-GUI L2, \modelname{}‑32B, 8B and 2B variants attain a 91.3\%, 88.8\%, 82.6\%, setting new state-of-the-art results at each scale. Additionally, the improvement on the high-level Advanced setting is much larger than on Basic setting, demonstrating strong grounding under high-level, implicit instructions.
\vspace{-1em}
\paragraph{Grounding for complex desktop scenario.} \textit{OSWorld-G} and \textit{OSWorld-G-Refine} assess grounding in complex desktop scenarios that require software commonsense, layout understanding, and fine-grained manipulation. As illustrated in Table~\ref{tab:osworld_g_comparison} and Table~\ref{tab:osworld_g_refine_comparison}, \modelname{} demonstrate consistent gains across categories and model sizes. On OSWorld-G, \modelname{}-32B achieves an average of 67.6\%, and increases to 70.9\% with zoom-in. This exceeds the strongest baselines, including UI-Venus-72B and GTA1-32B. \modelname{}-2B, and \modelname{}-8B also outperform baselines of similar scale by 5.0 and 5.4 points, respectively. Category-wise, the \modelname{} shows balanced performance, with the 32B variant achieving 78.5\% in Text Matching, 75.2\% in Element Recognition, 78.3\% in Layout Understanding, and 62.4\% in Fine-grained Manipulation. On OSWorld-G Refine, which reduces instruction ambiguity and emphasizes precise manipulation, \modelname{}-32B reaches 73.9\% and further improves to 75.0\% with zoom-in, exceeding strong baselines such as OpenCUA-32B and GTA1-32B.
\vspace{-1em}
\paragraph{Grounding across different operating systems.} \textit{ScreenSpot‑V2} spans mobile, desktop, and web interfaces with both text and icon grounding tasks. As shown in Table~\ref{tab:main_results_screenspot_v2_and_showdown} in Appendix, \modelname{}‑32B achieves a new state of the art with 96.5\% average accuracy, demonstrating strong results across all domains (e.g., 99.5\% on Desktop‑Text and 94.6\% on Web‑Icon). Notably, \modelname{}‑8B attains 95.5\% average accuracy, outperforming much larger models such as UI‑Venus‑72B and GTA1‑32B. The \modelname{}-2B on‑device variant reaches 92.5\%, surpassing many 7B baselines.
 
\begin{table*}[]
\caption{Performance comparison on \textbf{AndroidWorld} Benchmark. The best results are highlighted in \textbf{bold}, and the second-best results are \underline{underlined}.}
\label{tab:aw_com}
\centering
\footnotesize
\setlength{\tabcolsep}{20pt}
\renewcommand\arraystretch{1.2}
\begin{tabular}{lcc}
\toprule[1.2pt]
\textbf{\textsc{Model}}    & \textbf{\textsc{Paras.}}       & \textbf{\textsc{Success Rate}}  \\ 

\midrule
\rowcolor{gray!15}   
\multicolumn{3}{l}{\textit{Baselines}} \\
Qwen3-VL-2B~\citep{Qwen3-VL}                        &2B   & \cellcolor{light_purple} 36.4             \\
ScaleCUA-3B \citep{scale_cua}                        & 3B & \cellcolor{light_purple} 23.7 \\
Ferret-UI Lite-3B~\citep{Ferret-ui-lite}             & 3B & \cellcolor{light_purple} 28.0  \\
UI-Tars-7B~\citep{uitars}                            & 7B & \cellcolor{light_purple} 33.0   \\
UI-Tars-1.5-7B~\citep{ui-tars-15-seed}               & 7B & \cellcolor{light_purple} 30.0    \\
UI-Venus-7B~\citep{ui-venus}                         & 7B & \cellcolor{light_purple} 49.1     \\
GUI-Owl-7B~\citep{mobileagentv3}                     & 7B & \cellcolor{light_purple} 66.4      \\
Step-GUI-8B~\citep{step-gui}                     & 8B & \cellcolor{light_purple} 67.7      \\
Qwen3-VL-8B~\citep{Qwen3-VL}                        &8B   & \cellcolor{light_purple} 47.6              \\
Qwen3-VL-32B~\citep{Qwen3-VL}                       &32B  & \cellcolor{light_purple} 57.3               \\
UI-Tars-SFT-72B~\citep{uitars}                      & 72B & \cellcolor{light_purple} 46.6       \\

UI-Venus-72B~\citep{ui-venus}                       & 72B & \cellcolor{light_purple} 65.9        \\
Seed1.5-VL~\citep{guo2025seed15vl}                  &-  & \cellcolor{light_purple} 62.1           \\ 

Qwen3-VL-235B-A22B~\citep{Qwen3-VL}                 &235B & \cellcolor{light_purple} 63.7                \\ 
UI-Tars-1.5~\citep{ui-tars-15-seed}           & -   & \cellcolor{light_purple} 64.2         \\
Gemini-2.5-Pro~\citep{GeminiComputerUse}   &-    & \cellcolor{light_purple} 69.7            \\
Seed1.8 \citep{seed18} & - & \cellcolor{light_purple} 70.7   \\
UI-Tars-2~\citep{uitars2}                           &230B & \cellcolor{light_purple} \underline{73.3}          \\
\midrule
\rowcolor{gray!15}   
\multicolumn{3}{l}{Ours} \\
          
\modelname{}-2B                                      &2B & \cellcolor[HTML]{cfcdfd} 49.1~\textcolor{black}{\textbf{\scriptsize{}}} \\
          
\modelname{}-8B                                     &8B & \cellcolor[HTML]{cfcdfd} 70.7~\textcolor{black}{\textbf{\scriptsize{}}}\\
        
\modelname{}-32B                                    &32B & \cellcolor[HTML]{cfcdfd} \underline{73.3}~\textcolor{black}{\textbf{\scriptsize{}}} \\
\modelname{}-235B-A22B                              &235B & \cellcolor[HTML]{cfcdfd} \textbf{76.7}~\textcolor{black}{\textbf{\scriptsize{}}}  \\ \bottomrule[1.2pt]
\end{tabular}
    \vspace{-0.8em}
\end{table*}
\begin{wraptable}{r}{0.6\textwidth}
    \vspace{-10pt}
    \centering
    \small
    \caption{Performance comparison on Android Control (high-level instruction) and GUI Odyssey. Baseline results are mainly sourced from scores reported in \citep{zhang2025agentcpm}.} 
    \label{tab:ac_gui_odyssey}
    \begin{adjustbox}{max width=\linewidth}
        \begin{tabular}{l c c}
            \toprule
            \textbf{Model} &
            \textbf{AC-High} &
            \textbf{GUI-Odyssey} \\
            \midrule
            AgentCPM-GUI-8B~\citep{zhang2025agentcpm}      & 69.2 & 75.0 \\
            UI-TARS-7B~\citep{uitars}          & 74.4 & 67.9 \\
            OS-Atlas-7B~\citep{osaltas_and_screenspot_v2}          & 56.5 & \underline{76.8} \\
            Aguvis-7B~\citep{xu2024aguvis}            & 54.2 & 13.5 \\
            OdysseyAgent-7B~\citep{gui_odyssey}      & 32.7 & 73.7 \\
            UI-Venus-72B~\citep{ui-venus}         & \textbf{77.2} & 72.4 \\
            \midrule
            \rowcolor{gray!15}\textbf{\modelname{}-2B}        & 67.3 & 72.6  \\
            \rowcolor{gray!15}\textbf{\modelname{}-8B}    & 69.1 & 80.1 \\
            \rowcolor{gray!15}\textbf{\modelname{}-32B}         & \underline{75.5} & \textbf{83.4} \\
            \bottomrule
        \end{tabular}
    \end{adjustbox}
    \vspace{-10pt}
\end{wraptable}
\subsubsection{Mobile-Use Navigation Capability}
In addition to GUI grounding evaluation, we conduct a series of experiments to validate the effectiveness of mobile navigation capabilities. This evaluation comprises two components: static offline benchmarking and challenging dynamic online benchmarking.
\vspace{-0.5em}
\paragraph{Offline Benchmark}
We evaluate \modelname{} on Android Control~\citep{android_control} and GUI Odyssey~\citep{gui_odyssey}, two static offline benchmarks that assess GUI action execution, and grounding. Android Control comprises two instruction types: high-level instructions provide only a natural-language goal, whereas low-level instructions include step-wise action annotations. GUI Odyssey targets cross‑application navigation in mobile environments. As shown in Table~\ref{tab:ac_gui_odyssey}, \modelname{} produces competitive results on Android Control. On GUI Odyssey, \modelname{}‑32B achieves a new state of the art on the exact match score, substantially outperforming strong baselines and highlighting superior cross‑application navigation capability of our model.

\begin{table*}[]
\caption{Performance comparison on \textbf{MobileWorld} Benchmark (User-Int. is short for User-Interaction). The best results of end-to-end models are highlighted in \textbf{bold}, and the second-best results are \underline{underlined}.}
\label{tab:mobileworld_results}
\centering
\resizebox{\textwidth}{!}{
\begin{tabular}{lcccc}
\toprule[1.2pt]
\textbf{\textsc{Model}} & \textbf{\textsc{GUI-Only (116)}} & \textbf{\textsc{User-Int. (45)}} & \textbf{\textsc{MCP (40)}} & \textbf{\textsc{Overall}} \\ 
\midrule
\rowcolor{gray!15}   
\multicolumn{5}{l}{\textit{Agentic Framework}} \\
Claude-4.5-Sonnet \citep{claude_com} + UI-Ins-7B & 47.8 & 37.8 & 50.0 & \cellcolor{light_purple} 43.8 \\
Gemini-3-Pro \citep{gemini3pro} + UI-Ins-7B & 55.6 & 24.4 & 48.6 & \cellcolor{light_purple} 46.3 \\
GPT-5 \citep{gpt5} + UI-Ins-7B & 54.0 & 62.2 & 51.6 & \cellcolor{light_purple} 51.7 \\
\midrule
\rowcolor{gray!15}   
\multicolumn{5}{l}{\textit{End-to-End Model}} \\
GUI-Owl-7B \citep{mobileagentv3} & 7.7 & - & - & \cellcolor{light_purple} 4.5 \\
GUI-Owl-32B \citep{mobileagentv3} & 8.5 & - & - & \cellcolor{light_purple} 5.5 \\
UI-Venus-7B \citep{ui_venus} & 8.5 & 2.3 & - & \cellcolor{light_purple} 5.5 \\
UI-Venus-72B \citep{ui_venus} & 16.4 & 4.7 & - & \cellcolor{light_purple} 10.4 \\
Qwen3-VL-8B \citep{Qwen3-VL} & 9.4 & 0.0 & 0.0 & \cellcolor{light_purple} 5.5 \\
Qwen3-VL-32B \citep{Qwen3-VL} & 11.9 & 6.7 & 2.7 & \cellcolor{light_purple} 9.0 \\
Qwen3-VL-235B-A22B \citep{Qwen3-VL} & 12.8 & 4.4 & 5.4 & \cellcolor{light_purple} 9.5 \\
Doubao-1.5-UI-TARS \citep{ui-tars-15} & 26.3 & 32.4 & - & \cellcolor{light_purple} 20.9 \\
\midrule
\rowcolor{gray!15}   
\multicolumn{5}{l}{\textit{Ours}} \\
\modelname{}-8B &  27.5 &  22.2 &  20.0 & \cellcolor[HTML]{cfcdfd} 24.9 \\
\modelname{}-32B & \underline{36.2} & \underline{46.7} & \underline{30.0} & \cellcolor[HTML]{cfcdfd} \underline{37.3} \\
\modelname{}-235B-A22B & \textbf{39.7} &  \textbf{51.1} &  \textbf{37.5} & \cellcolor[HTML]{cfcdfd} \textbf{41.7} \\
\bottomrule[1.2pt]
\end{tabular}
}  
    \vspace{-0.8em}
\end{table*}
\vspace{-0.5em}
\paragraph{Online Benchmark}
Online evaluation provides a more realistic assessment of agent capability than static offline tests, as it requires multi-turn adaptive perception, reasoning, and action in dynamic environments.
We evaluate \modelname{} on online benchmark AndroidWorld ~\citep{android_world}, a live emulator–based benchmark comprising 116 tasks across 20 mobile applications. For a fair comparison, we report results only for end-to-end GUI agents, isolating the intrinsic GUI capability of foundation models without interference from pipeline frameworks or external tools.

As shown in Table~\ref{tab:aw_com}, \modelname{} achieves new state-of-the-art results on AndroidWorld across model scales. \modelname{}‑235B‑A22B attains a 76.7\% success rate, surpassing UI‑TARS‑2 (73.3\%) and Gemini‑2.5‑Pro (69.7\%) by +3.4 and +7.0 points, respectively. The 32B variant reaches 73.3\%, establishing SOTA at its scale and outperforming larger baseline models, including UI‑Venus‑72B, Gemini‑2.5‑Pro, and UI‑TARS-1.5. \modelname{}‑8B achieves 70.7\% sucess rate, exceeding strong similar‑scale models such as GUI‑Owl‑7B and Step-GUI-8B. Finally, the 2B model attains 49.1\% success rate, improving over the strongest on‑device baseline Ferret‑UI‑Lite‑3B (28.0\%), by +21.1 points (+75.4\% relative), providing a strong foundation for our native device-cloud collaboration system. These results demonstrate the strong GUI capability of \modelname{} across the full spectrum of scales.
\vspace{-0.5em}
\paragraph{Real-world-oriented online evaluation.} Beyond AndroidWorld, we assess our models on MobileWorld \citep{mobile_world}, a more challenging and more realistic online benchmark with 201 tasks across 20 applications. MobileWorld emphasizes long‑horizon, cross‑application tasks and includes evaluation beyond pure UI operation, such as MCP tool use and agent–user interaction. As summarized in Table~\ref{tab:mobileworld_results}, \modelname{} achieve substantial improvements against end‑to‑end models: \modelname{}‑235B‑A22B reaches 41.7\% overall success rate, \modelname{}‑32B attains 37.3\%, and \modelname{}‑8B scores 24.9\%. These outperform the strongest end‑to‑end baseline (Doubao-1.5-UI-TARS, 20.9\%) by +20.8, +16.4, and +4.0 points, respectively.\\
We also compare against agentic frameworks that use a strong planner such as GPT-5 and an external grounding model (UI-Ins \citep{uiins}) as executor. These pipelines benefit from strong planning and reasoning ability and an external execution model, making the comparison not strictly fair. However, our pure end‑to‑end models remain competitive, with \modelname{}‑235B‑A22B reaching 41.7\% overall success rate, close to Gemini‑3‑Pro+UI‑Ins (46.3\%) and Claude‑4.5‑Sonnet+UI‑Ins (43.8\%).
\begin{figure}
    \centering  
        \begin{subfigure}[t]{0.99\textwidth}
            \centering
            \includegraphics[width=0.99\linewidth]{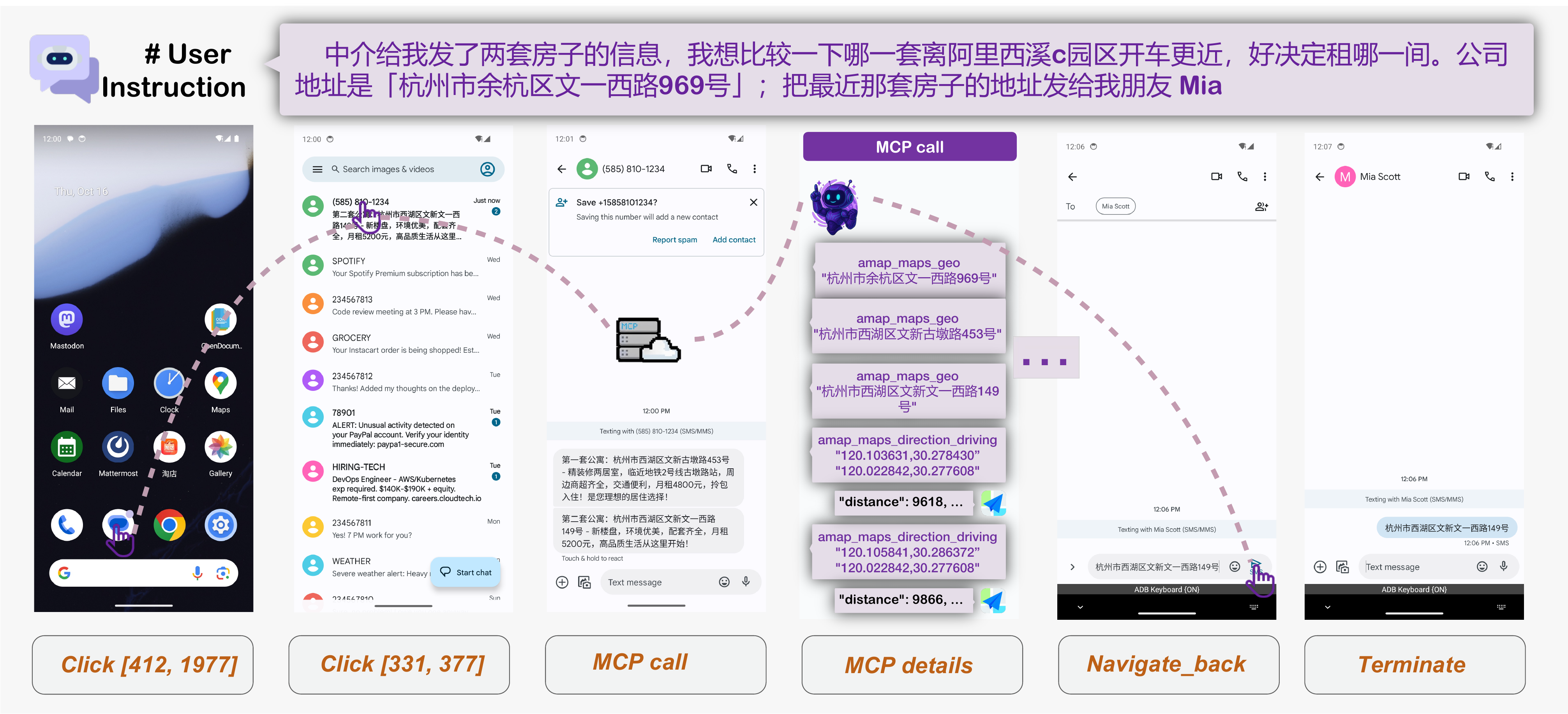}
            \caption{}
            \label{fig:mcp_1}
        \end{subfigure}
        \begin{subfigure}[t]{0.99\textwidth}
            \centering
            \includegraphics[width=0.99\linewidth]{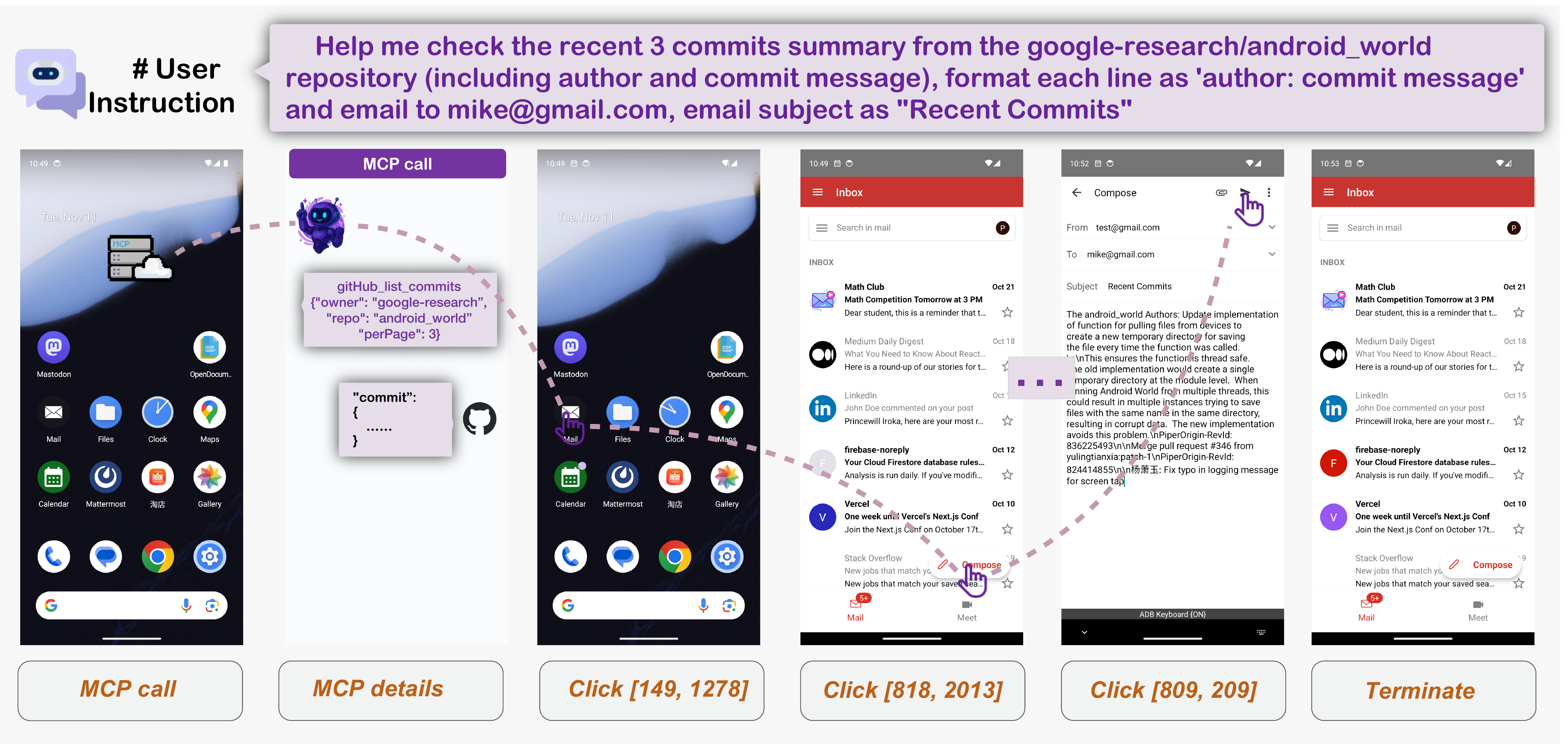}
            \caption{}
            \label{fig:mcp_2}
        \end{subfigure}
    \caption{Case studies of MCP tool using of \modelname{}. (a): Using MCP tools provide shortcuts that compress multiple UI actions into a few API calls; (b): Using MCP tools brings traditionally desktop-only workflows (e.g., GitHub commit search) to mobile. The user instruction for (a) is: \textit{``Compare the two apartment listings sent by the agent and determine which has the shorter driving time to Alibaba Xixi Campus (Zone C; 969 Wenyi West Road, Yuhang District, Hangzhou). Send the address of the nearer apartment to my friend Mia''.}}
    \label{fig:mcp_case}
        \vspace{-0.8em}
\end{figure}

To quantitatively assess our model’s MCP tool-use and agent–user interaction capabilities, we evaluate on the MobileWorld benchmark’s two relevant subsets (User-Int., 45 tasks; MCP, 40 tasks). Our \modelname{}-235B-A22B achieves 51.1\% on User-Int. and 37.5\% on MCP, outperforming existing end-to-end baselines by +18.7 and +32.1 points, respectively (best prior scores: 32.4\% and 5.4\%). Compared with agentic frameworks, our end-to-end model is competitive on agent–user interaction, surpassing Gemini-3-Pro+UI-Ins-7B (24.4\%) and Claude-4.5-Sonnet+UI-Ins-7B (37.8\%). Overall, these results demonstrate strong MCP tool-use and agent–user interaction capabilities relative to end-to-end models, and competitive performance compared with agentic frameworks.

\begin{figure}
    \centering  \includegraphics[width=0.99\linewidth]{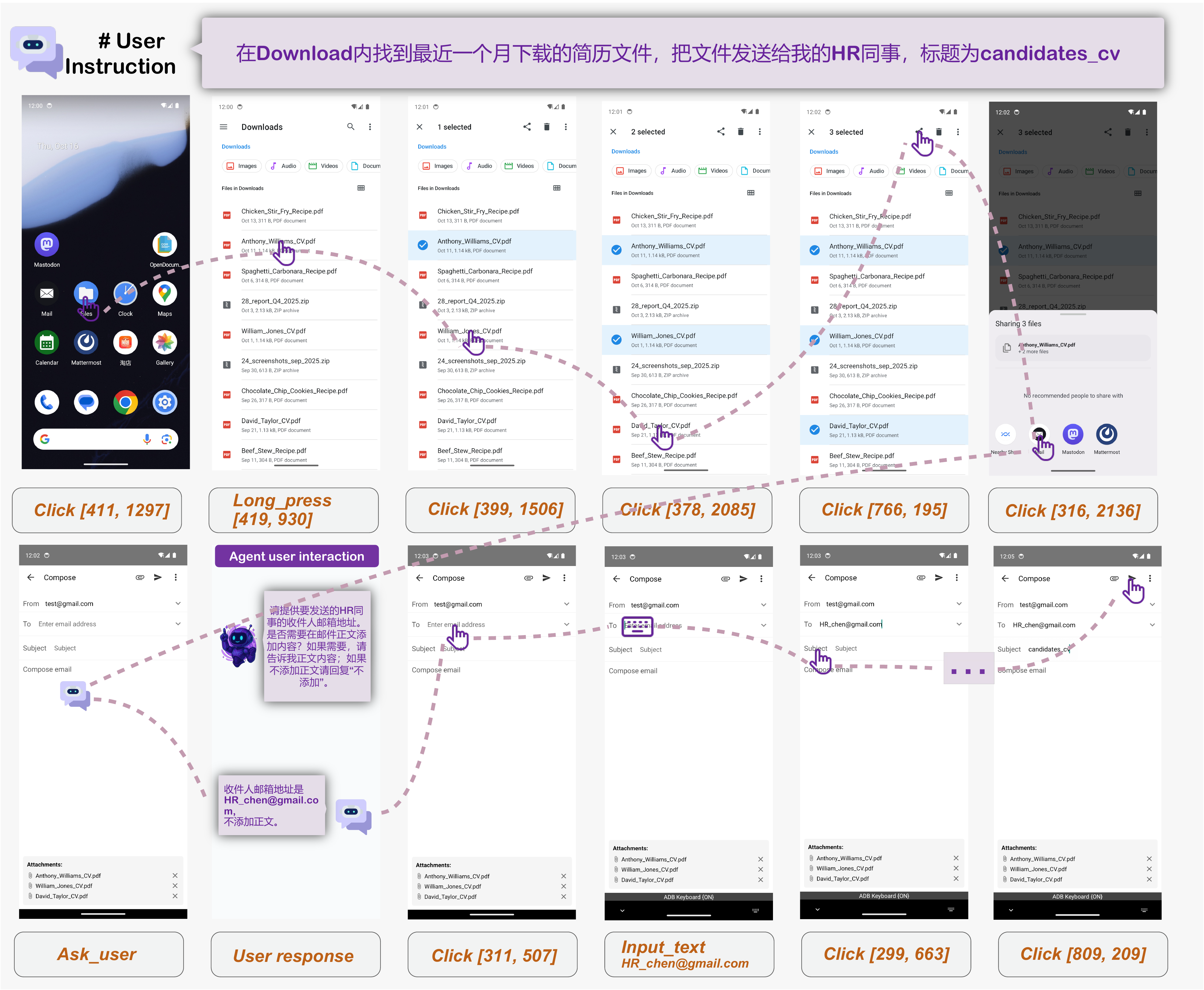}
    \caption{A case study of agent user interaction. The user instruction is: \textit{``In the Downloads folder, locate resume file(s) downloaded within one month and send them to my HR colleague with the subject “candidates\_cv.''}.}
    \label{fig:call_user}
        \vspace{-0.3em}
\end{figure}
\subsection{MCP Augmentation and Agent-User Interaction}
\label{mcp_call_user}
In addition to the quantitative results, we present case studies for the MCP tool use and user-agent interaction capability of \modelname{}.
\vspace{-0.5em}
\paragraph{MCP-Augmented Tasks}
Figure \ref{fig:mcp_case} shows two representative scenarios that benefit from MCP tool use, illustrating two core benefits: compressing multi step GUI operations into a few tool calls and enabling traditionally desktop only workflows on mobile. \\
Figure \ref{fig:mcp_1} illustrates a realistic cross-application task that aggregates information and compares route distance. The user receives two apartment addresses by SMS and asks the agent to compare driving time from Alibaba Xixi Campus, then send the nearer address to Mia. Traditionally, this would require repeated switching between SMS and a maps app, copying and pasting addresses, and running two separate route searches. With an Amap MCP call, the agent can simply sets the campus as the origin and each candidate address from the SMS as the destination, and retrieves structured travel time and distance. This process compresses multiple GUI operations into a small number of tool calls, significantly reducing GUI interactions and improving end-to-end efficiency. \\
\begin{figure}
    \centering
    \includegraphics[width=1.0\linewidth]{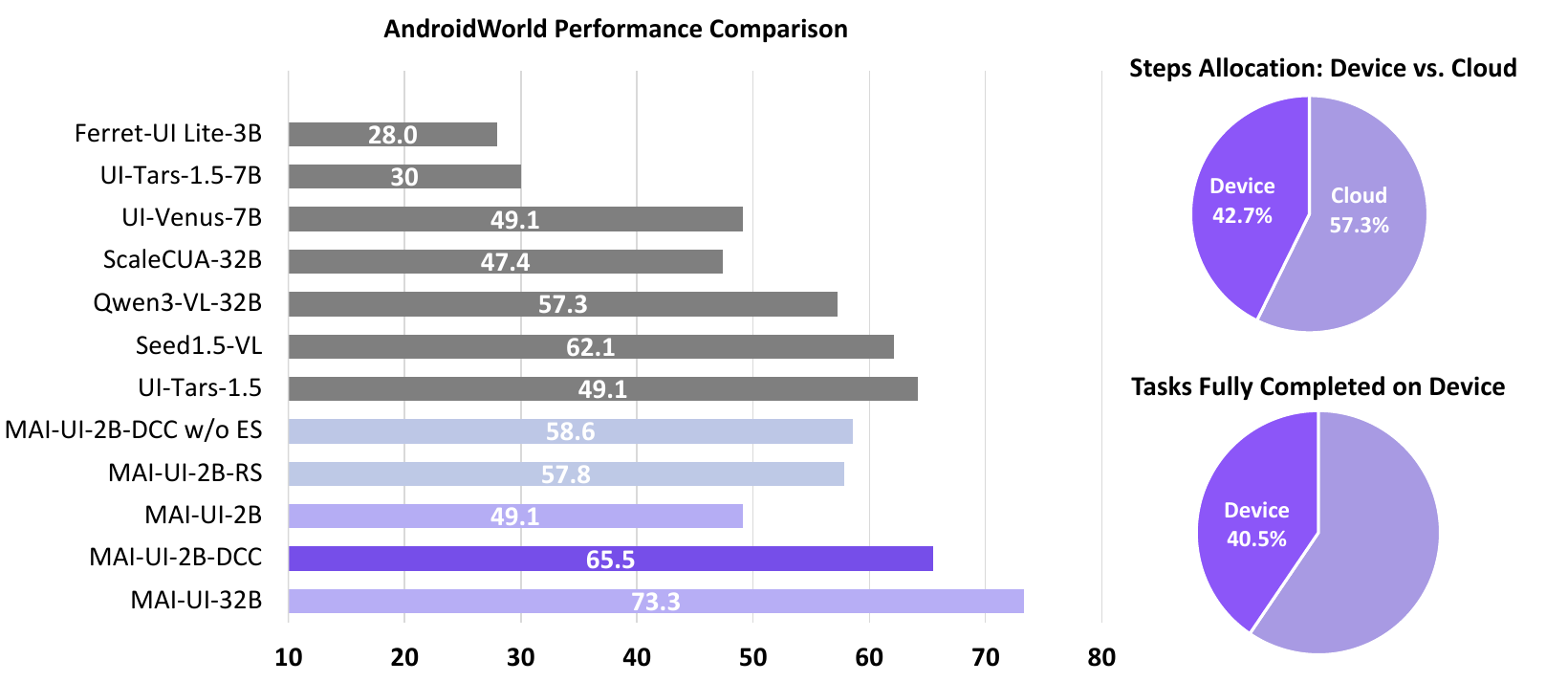}
    \caption{The native device–cloud collaboration (DCC) capability significantly improves the on-device model’s online performance, surpassing the random switch (RS) baseline, w/o error summary (ES) baseline, and several larger pure‑cloud models by a substantial margin. Our native DCC system also improves efficiency, executing 42.7\% of steps locally and completing 40.5\% of tasks entirely on‑device, thereby reducing cloud calls.}
    \label{fig:device_cloud_exp}
        \vspace{-0.5em}
\end{figure}
\begin{figure}[h]
    \centering  \includegraphics[width=0.99\linewidth]{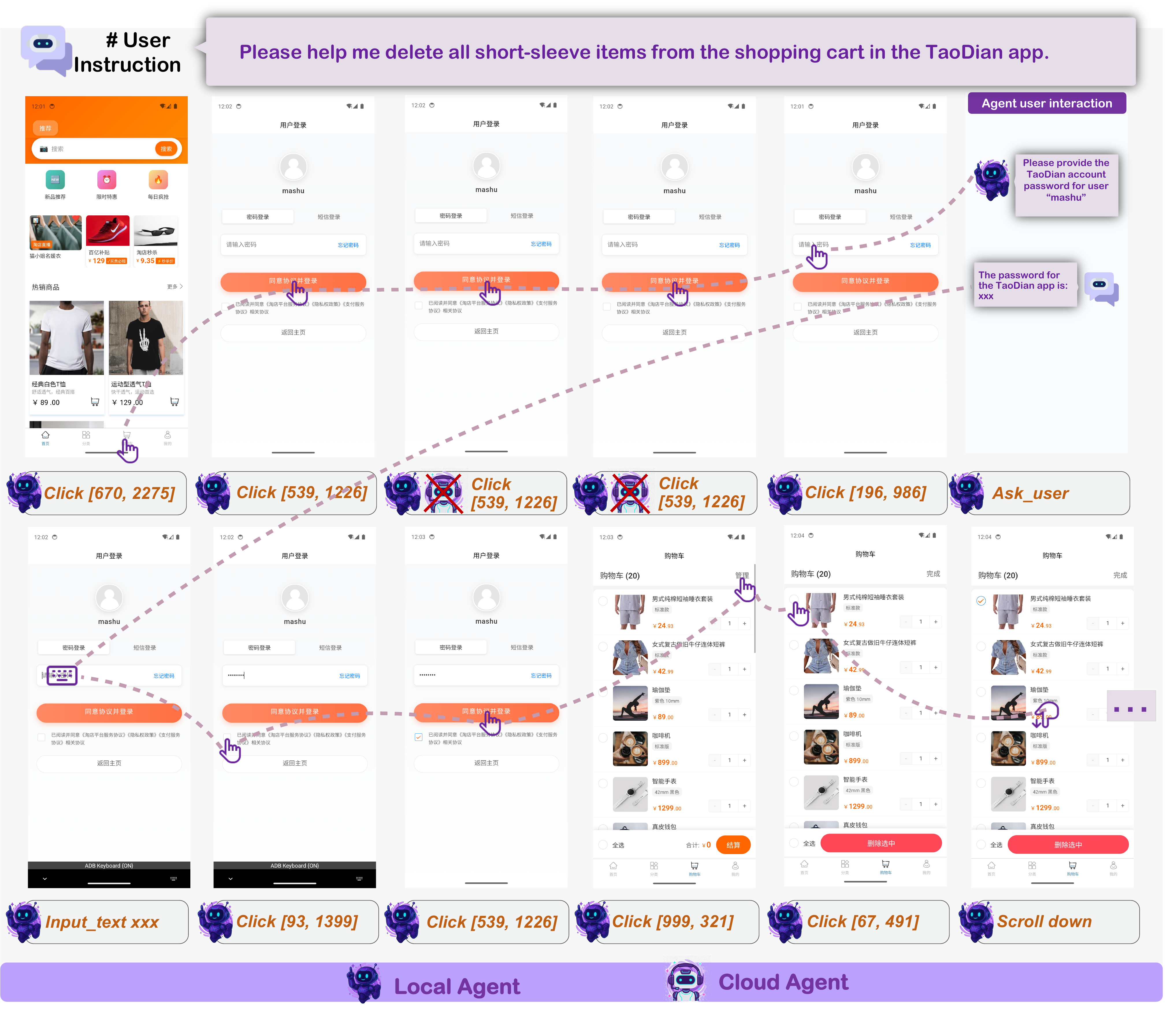}
    \caption{A pilot study demonstrating privacy protection in the device-cloud collaboration system. At step 3, the local agent deviates, as it repeatedly tapping the Login button without entering a password. The trajectory monitor flags the misalignment and proposes switch to the cloud agent. However, the privacy detection module detects sensitive credentials and blocks the switch, keeping execution on device. The local monitor ultimately corrects the trajectory and completes the task.}
    \label{fig:privacy}
        \vspace{-0.5em}
\end{figure}
Figure \ref{fig:mcp_2} shows a workflow that is usually handled on desktop environments. The user asks the agent to retrieve the author and commit message from an GitHub repository, and send the summary by email. On mobile, browsing commit history is inconvenient due to limited screen space. Through MCP API calls, the agent directly queries the repository and receives commit metadata in structured form, and extracts the fields required by the task. This case demonstrates that using MCP tools can not only compress UI operations but also expand the capability of mobile GUI agent, providing access to services that are commonly available in desktop applications.
\vspace{-0.5em}
\paragraph{Agent-User-Interaction}
To evaluate the agent’s ability to interact with the user when necessary, we present a case study of a file-sharing task that requires proactive clarification (Figure \ref{fig:call_user}). The user instructs the agent to locate recent resume files in the Downloads folder and send them to an HR colleague, but several critical parameters are under-specified, including the recipient’s email address and the email body. Detecting these gaps, \modelname{} pauses execution and issues an \texttt{ask\_user} action to request the missing details. After receiving the user’s response, the agent resumes the GUI trajectory: it auto-fills and sends the email. This example demonstrates \modelname’s proactive agent–user interaction capability, which is essential in realistic GUI tasks where instructions are often ambiguous or incomplete. Overall, the combination of MCP enhancement and Agent-User Interaction enables the model to better handle real-world tasks.
\subsection{Device-Cloud Collaboration Analysis}
\label{device_cloud_exp_sec}
\paragraph{On-device capability gains.} On the AndroidWorld online benchmark, we evaluate how device–cloud collaboration (DCC) enhances the on-device model’s performance. Our system uses \modelname{}‑2B as the local agent and \modelname{}‑32B as the cloud model. We compare our DCC system with four settings: (i) Local GUI agent, which uses the on-device model solely as the GUI agent; (ii) Cloud GUI agent, which relies on large scale cloud GUI agents; and (iii) Random switch (RS) baseline, which calls the cloud model at the same frequency as our DCC method but without monitor-guided switch. As shown in the left part of Figure \ref{fig:device_cloud_exp}, our DCC system achieves relative improvement of {$\mathbf{33.4\%}$} over the on-device model, significantly strengthening its ability to complete mobile tasks. It further surpasses the RS baseline by $7.7\%$, demonstrating that monitor-driven cloud-model switching effectively triggers cloud assistance and boosts collaboration quality. Finally, compared with several pure cloud GUI agent baselines, our device–cloud system achieves higher scores, demonstrating strong performance under realistic deployment setting.
\vspace{-1em}
\paragraph{Efficiency gains versus cloud-only solution.} We analyze cloud usage and on-device completion under our device–cloud collaboration system in the right part of Figure \ref{fig:device_cloud_exp}. Compared to cloud-only serving, our system reduces cloud model calls by {42.7\%}, substantially lowering serving cost and latency for using cloud model. On AndroidWorld, over $40\%$ of tasks are completed entirely on device, further confirming that unnecessary cloud model calls are reduced. These results demonstrate that our device–cloud collaboration obtains substantial efficiency gains compared to cloud-only solutions.
\vspace{-1em}
\paragraph{Impact of error summary information.} To assess the effectiveness of monitor-generated error summaries, we run an ablation that removes the error summary (ES) at switching and compare against our original system. As shown in the left part of Figure \ref{fig:device_cloud_exp}, compared to the baseline that without error summary, providing the error summary yields a +6.9 increase in task success rate, underscoring its importance for the cloud agent's trajectory recovery process.
\vspace{-1em}
\paragraph{Privacy preservation.} We present a pilot case study to showcase the system’s privacy protection capability. We introduce an additional local privacy monitor, which blocks cloud switch whenever privacy‑sensitive content is present, even if a trajectory deviation is detected. As shown in Figure \ref{fig:privacy}, for a task involving user‑sensitive content (password entry), the system continues on‑device execution even when the monitor detects trajectory deviation (repeated clicks on the login icon in steps 2–4), thereby adhering privacy constraints. The Local Agent ultimately corrects the trajectory and completes the task. Throughout this case, no privacy‑sensitive content is transmitted to the cloud, demonstrating that this device-cloud collaboration system effectively protects user privacy.

\begin{wraptable}{r}{0.49\textwidth}
    \centering
        \vspace{-10pt}
    \caption{Online RL performance gains and ablations on standard GRPO and interaction budget per trajectory (max\_env\_steps)}
    \begin{adjustbox}{max width=\linewidth}
        \small
        \begin{tabular}{l c}
        \toprule
        \textbf{Method} & \textbf{AndroidWorld} \\
        \midrule
        \textit{SFT vs. RL Comparison} & \\
        \midrule
        MAI-UI-2B-SFT & 45.1 \\
        MAI-UI-2B-RL & \textbf{49.1}\,(+4.0) \\
        \addlinespace[0.3em]
        MAI-UI-8B-SFT & 64.7 \\
        MAI-UI-8B-RL & \textbf{70.7}\,(+6.0) \\
        \addlinespace[0.3em]
        MAI-UI-32B-SFT & 69.8 \\
        MAI-UI-32B-RL & \textbf{73.3}\,(+3.5) \\
        \midrule
        \textit{RL Ablation on 8B} & \\
        \midrule
        MAI-UI-8B-SFT & 64.7 \\
        GRPO (max\_env\_steps=50) & 66.5\,(+1.8) \\
        Our method (max\_env\_steps=15) & 66.4\,(+1.7) \\
        Our method (max\_env\_steps=30) & 68.5\,(+3.8) \\
        Our method (max\_env\_steps=50) & \textbf{70.7}\,(+6.0) \\
        \bottomrule
        \end{tabular}
        \label{tab:rl_ablation}
        \end{adjustbox}
    \vspace{-10pt}
\end{wraptable}
\subsection{Online RL Analysis}
\label{online_rl_exp_sec}

\begin{figure}[!t]
    \centering
    \begin{subfigure}[t]{0.45\textwidth}
        \centering
        \includegraphics[width=\textwidth]{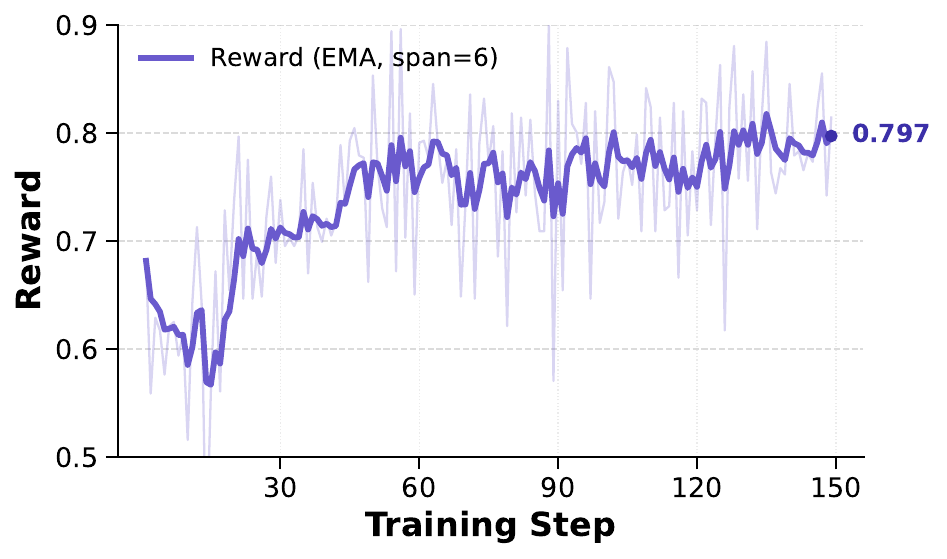}
        \vspace{-1.5em}
        \caption{}
        \label{fig:reward}
    \end{subfigure}
    \hfill
    \begin{subfigure}[t]{0.45\textwidth}
        \centering
        \includegraphics[width=\textwidth]{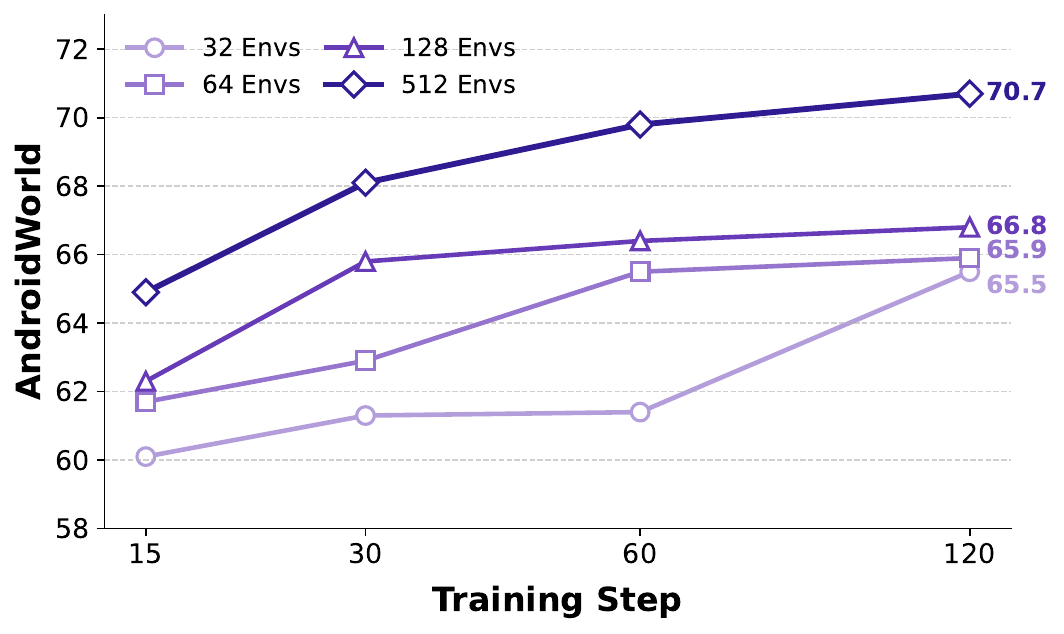}
        \vspace{-1.5em}
        \caption{}
         \label{fig:android_world_perf}
    \end{subfigure}
    
    \caption{MAI-UI-8B-RL training details: (a) train set reward trend (b) performance scaling with the number of environments.}
    \label{fig:rl_details}
        \vspace{-0.8em}
\end{figure}

\paragraph{Performance Gains.}
As shown in Table \ref{tab:rl_ablation}, online RL consistently enhances performance across all model scales. Specifically, the 2B model achieves an absolute improvement of \textbf{4.0 percentage points} (45.1\% $\rightarrow$ 49.1\%), the 8B model gains \textbf{6.0 percentage points} (64.7\% $\rightarrow$ 70.7\%), and the 32B model improves by \textbf{3.5 percentage points} (69.8\% $\rightarrow$ 73.3\%). These results correspond to relative improvements of 8.9\%, 9.3\%, and 5.0\%, respectively, demonstrating that online RL effectively enhances agent performance regardless of model size. Figure \ref{fig:reward} further illustrates this improvement, showing that the reward metric steadily increases throughout the training process.

\begin{figure}[t!]
    \centering
    \includegraphics[width=0.99\linewidth]{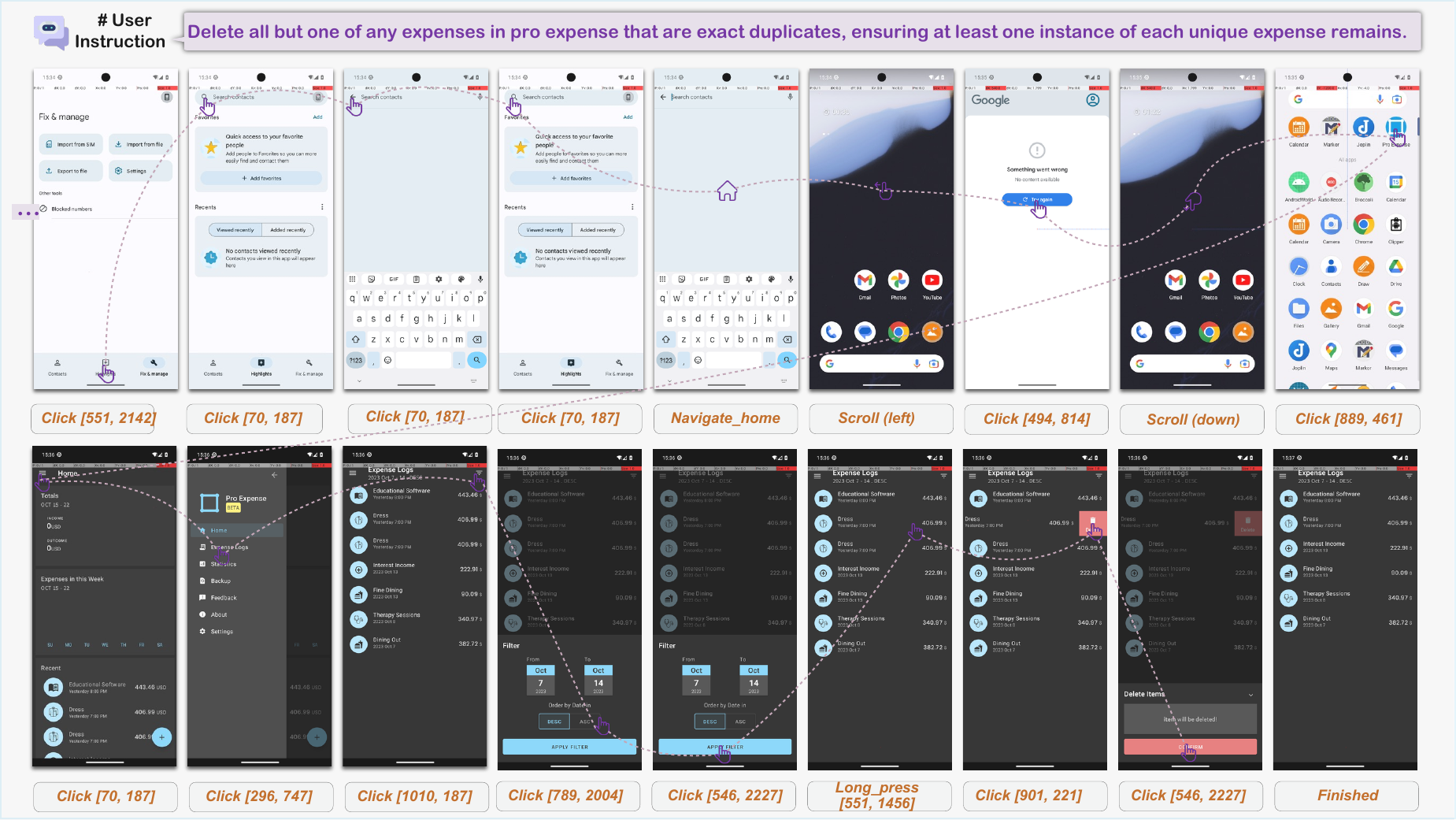}
    \caption{\modelname{} shows robustness in recovering from recovering from failures: the agent initially navigates to the wrong application, but \modelname{} correct the trajectory and complete the task.}
    \label{fig:error_recover}
        \vspace{-0.5em}
\end{figure}

\vspace{-0.5em}
\paragraph{Experimental Analysis.} We perform ablation studies to evaluate the proposed key components in online RL, with the results shown in Table \ref{tab:rl_ablation} and Figure \ref{fig:rl_details}.
\vspace{-0.5em}
\begin{itemize}[leftmargin=*]
    \item \textbf{Comprison with standard GRPO.} Standard GRPO applied after SFT yields a modest gain of +1.8 percentage points on AndroidWorld. In contrast, our enhanced GRPO with data curriculum, repetition penalty, and experience replay achieves a +6.0 percentage point improvement, delivering an additional +4.2 percentage points over the baseline.
    \item \textbf{Effect of interaction budget.} The maximum environment interaction budget per trajectory substantially influences performance. Extending the budget from 15 to 30 and subsequently to 50 steps yields progressive improvements (+1.7, +3.8, and +6.0 percentage points, respectively). A larger budget enables more extensive rollouts and provides richer exploration opportunities during training. 
    \item \textbf{Impact of image resolution.} Image resolution critically affects online RL efficiency, as higher resolutions introduce more visual tokens and slow down both training and inference. Leveraging the relative coordinate mechanism of Qwen3-VL, we found that 720p resolution achieves performance comparable to 1080p while providing a $\sim$50.1\% speedup per step. Conversely, 540p resolution, despite faster processing, substantially degrades model performance due to insufficient detail for fine-grained UI element perception.
\item \textbf{Scaling parallel environments.} Figure~\ref{fig:android_world_perf} shows how parallel environment count affects model performance. Increasing parallel environments from 32 to 512 significantly accelerates learning and improves final performance (65.5\% $\rightarrow$ 70.7\%). Training with fewer environments exhibits early saturation, indicating that limited environments constrain policy improvement. These findings highlight that scaling parallel environments to enhance exploration diversity is critical for overcoming performance bottlenecks in GUI agent RL training.
\end{itemize}
\begin{figure}[!t]
    \centering
    \begin{subfigure}[t]{0.45\textwidth}
        \centering
        \includegraphics[width=\textwidth]{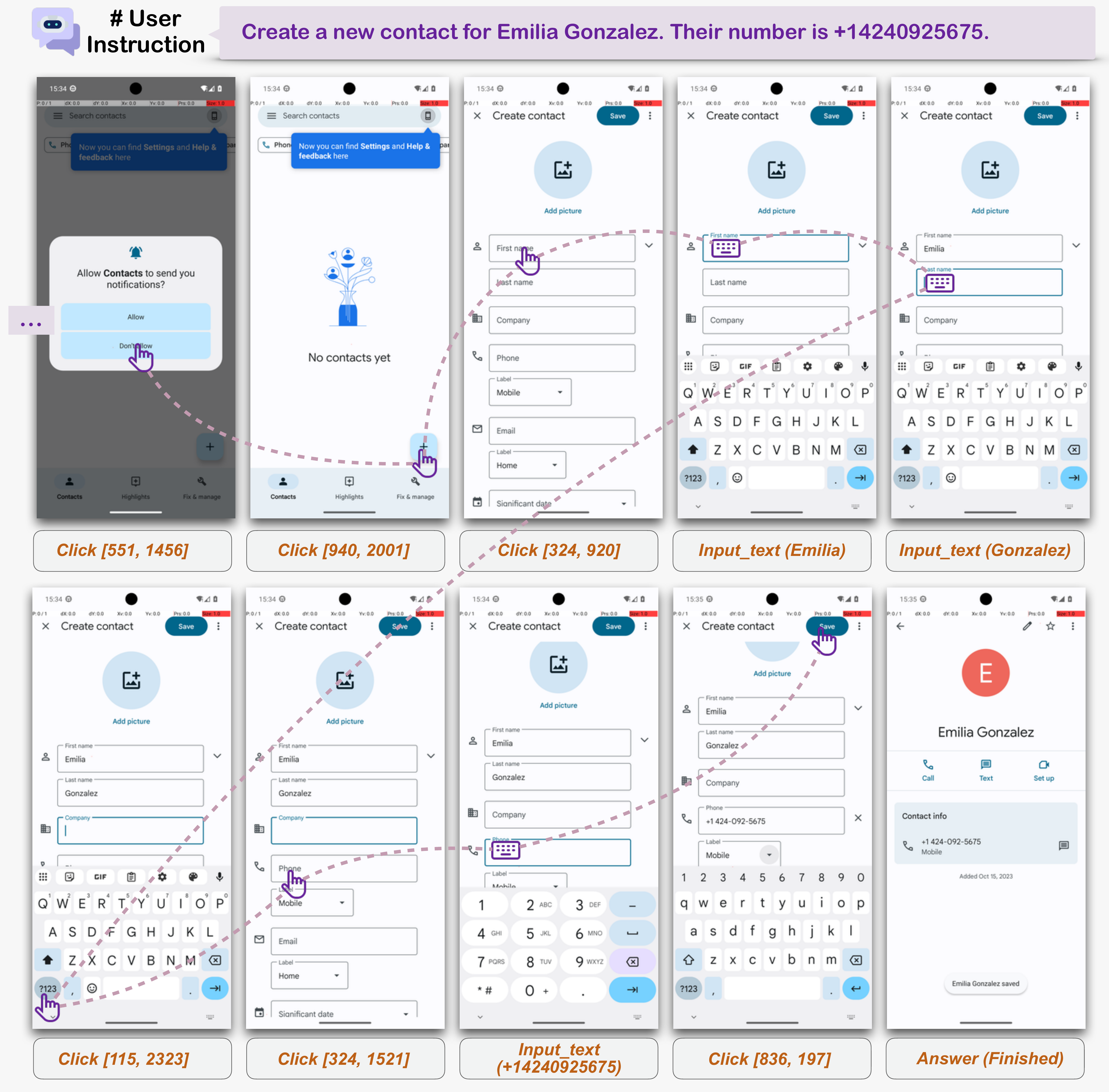}
        \vspace{-1.5em}
        \caption{}
        \label{fig:robustness_permission}
    \end{subfigure}
    \hfill
    \begin{subfigure}[t]{0.54\textwidth}
        \centering
        \includegraphics[width=\textwidth]{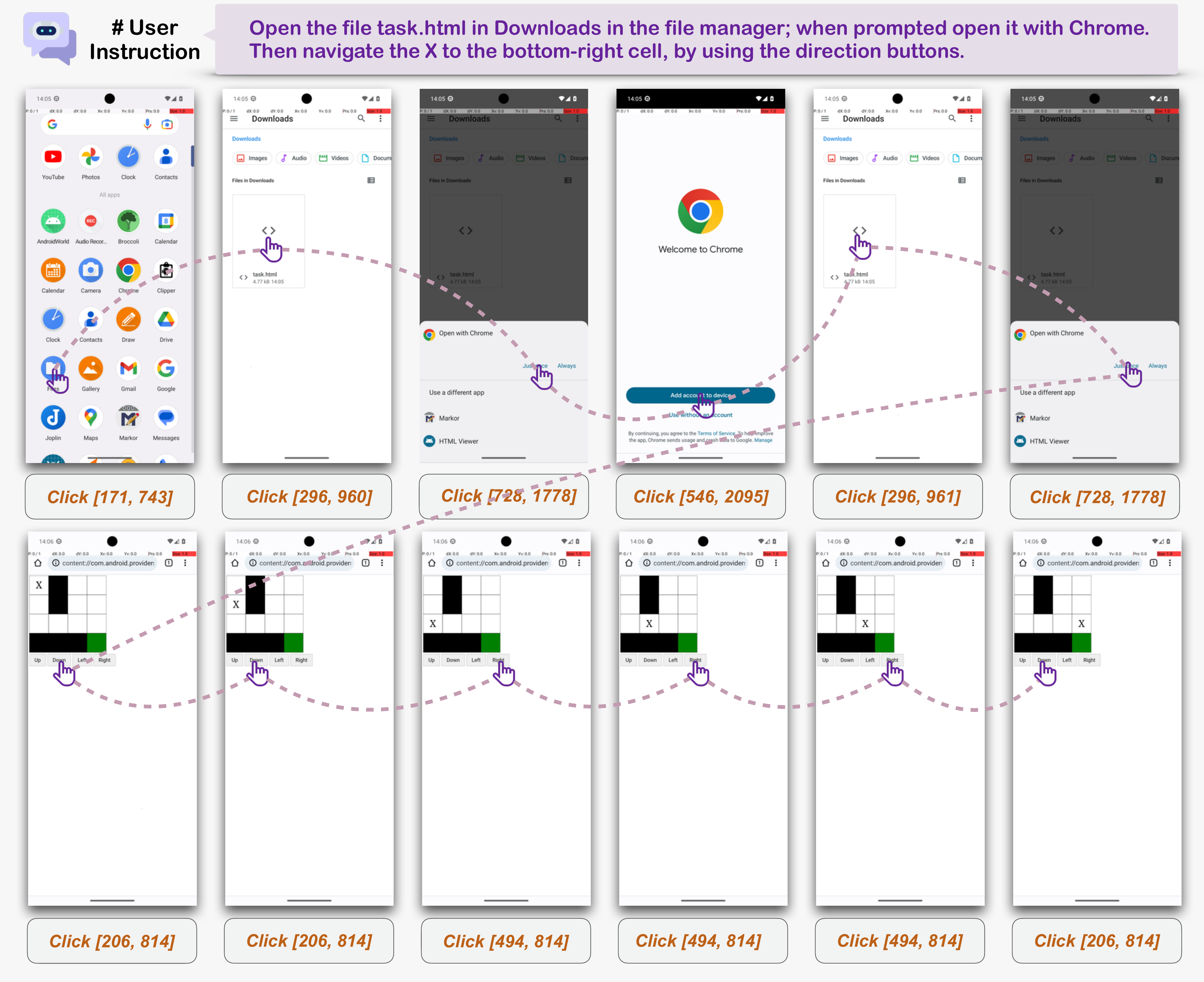}
        \vspace{-1.5em}
        \caption{}
        \label{fig:robustness_popup}
    \end{subfigure}
    
    \caption{\modelname{} demonstrates robustness to unexpected permission dialogs (a) and pop-ups (b) in AndroidWorld, and reliably resumes the task.}
    \label{fig:robustness}
        \vspace{-0.4em}
\end{figure}
\vspace{-1em}
\paragraph{Case Studies: Enhanced Robustness.}
Figures~\ref{fig:robustness} and~\ref{fig:error_recover} illustrate qualitative improvements in robustness after online RL training. Figure~\ref{fig:robustness} demonstrates \modelname{}'s ability to handle unexpected permission dialogs and pop-ups during task execution. When creating a new contact for "Emilia Gonzalez," the agent encounters a notification permission request that was not present during offline training. The RL-trained model successfully dismisses the dialog and continues task execution without deviation, whereas the base model often fails to recover from such interruptions. Figure~\ref{fig:error_recover} showcases the agent's capability to recover from failed actions in a complex expense management task. When instructed to delete duplicate expenses, the agent initially navigates to the wrong application. Nevertheless \modelname{} correct the trajectory and compleete the task in the following steps. These case studies highlight that online RL training substantially improves the agent's robustness to real-world unpredictability. This is difficult to acquire through offline training alone, where the diversity of edge cases and failure modes is inherently limited.
\subsection{Analysis on Grounding}
\label{sec:grounding_analysis}
As demonstrated by our prior work in UI-Ins~\citep{uiins}, our Instruction-as-Reasoning approach offers benefits from several perspectives:
\vspace{-0.5em}
\begin{itemize}[leftmargin=*]
\item \textbf{Reasoning that helps grounding.} Prior studies find that general chain-of-thought reasoning can often degrade grounding performance \citep{ui_r1, GTA1, GUIG1}. Inspired by how humans approach grounding, we train models to use diverse instruction perspectives as explicit reasoning pathways, making reasoning actionable and beneficial for GUI grounding.
\item \textbf{Mitigating policy collapse in SFT + RL framework.} Policy collapse often occurs in grounding after SFT with coordinate-only supervision \citep{phi_ground}. Instruction-as-Reasoning stabilizes RL by pretraining the model to generate diverse reasoning pathways, which enhances exploratory behavior and stabilizes policy optimization in the RL phase.
\item \textbf{Emergent multi-perspective capabilities.} After employ our instruction-as-reasoning method, the model can strategically select appropriate reasoning perspectives given different contexts and compose multiple perspectives into a cohesive one. Interestingly, it can also generate novel analytical angles beyond the four trained perspectives.
\end{itemize}

\section{Related Works}
\subsection{GUI Grounding}
GUI grounding is the foundational capability of GUI agents that maps natural language instructions to the locations of target elements in screenshots. Prior GUI grounding methods mainly focus on training in a Supervised Fine-Tuning (SFT) paradigm, such as JEDI~\citep{jedi}, OS-Atlas~\citep{osaltas_and_screenspot_v2}, Aguvis~\citep{aguvis}, Uground~\citep{uground} and Aria-UI~\citep{ariaui}. Reinforcement learning methods, particularly GRPO~\citep{grpo} have demonstrated remarkable sucess on various visual-language tasks, including Semantic Segmentation~\citep{segzero}, Visual Question-Answering~\citep{visionreasoner,vision-r1} and Temporal Video Grounding~\citep{time-r1}. Consequently, recent efforts have increasingly focused on adapting RL for GUI grounding. GUI Grounding methods like GUI-R1~\citep{gui_r1}, GUI-Actor~\citep{gui_actor} and GTA1~\citep{GTA1} play as an pioneer role in pure RL paradigm and surpass SFT-based methods by a large margin. However, a key limitation of a pure RL paradigm is that it overlooks the substantial benefit offered by an initial SFT stage. While InfiGUI-R1~\citep{infiguir1} achieved success with an SFT+RL framework by reframing GUI grounding as a trajectory-level task that encourages model reflection, the  SFT+RL paradigm remains difficult to implement in practice. It is also demonstrated by Phi-Ground~\citep{phi_ground} that SFT+RL framework is prone to policy collapse. Our grounding method overcomes this issue by using the SFT stage to teach the model diverse reasoning through different instruction perspectives, and then utilizes the RL stage to further incentivize the model to select the appropriate reasoning pathway, thereby establishing an effective example for the SFT+RL training paradigm.
\subsection{GUI Navigation}
Moving beyond single-step grounding, research on GUI agents has advanced to GUI navigation tasks, where agents must execute multi-step action sequences to complete a user instruction. Early works achieve this goal through agent frameworks that contain multiple components, such as a planner and a grounding module \citep{mobile-agent, ariaui, agents2}. For practical deployment and cost efficiency, recent efforts increasingly target unified vision–language–action models that jointly learn grounding and navigation \citep{ui-tars, zhang2025agentcpm, ui_venus}. Most of these models are large-scale to enable stronger reasoning and planning, while a growing line of work builds competitive small on-device models for lower latency, improved privacy, and resilience to connectivity constraints \citep{Ferret-ui-lite}. To meet deployment needs, we release both large-scale and on-device models, and further integrate a native device–cloud collaboration capability.\\
With the rapid development in GUI navigation, progress has also been made in real-world deployment of GUI agents \citep{Gemini-Live}. In practice, however, current agents function as copilots rather than standalone executors, making effective agent–user interaction a critical yet often overlooked capability. Moreover, most agents remain limited to pure UI manipulation, and only recently have efforts begun to equip them with extended SDK functions \citep{uitars2}. Integrating external tools such as the MCP can compress long, brittle UI action sequences into a handful of API calls and unlock desktop workflows that were previously infeasible on mobile.

\section{Conclusion}
In this work, we present \modelname{}, a family of foundation GUI agents. To address challenges for real-world deployment of GUI agents, we introduce three main components: a self-evolving data pipeline that expands the navigation data to include user interaction and MCP tool calls, a native device–cloud collaboration system that routes execution by task state and data sensitivity, and an online RL framework with advanced system optimizations. Experimental results show that \modelname{} sets a new state of the art on grounding benchmarks and both online and offline evaluation of mobile use navigation. Benchmark results on MobileWorld and qualitative case studies verify effective agent–user interaction and MCP-enabled tool use capability of \modelname{}. The device–cloud collaboration substantially improves on-device performance while reducing cloud model calls, yielding performance, privacy and cost benefits. Taken together, these advances make \modelname{} a step further toward practical foundation GUI agent for mobile use.

\clearpage
\bibliography{biblio}
\bibliographystyle{colm2024_conference}

\clearpage
\appendix
\section{Additional Grounding Results}
We present more grounding results of ScreenSpot-v2 and OSWorld-G Refine in Table \ref{tab:main_results_screenspot_v2_and_showdown}, and Table \ref{tab:osworld_g_refine_comparison}.
\begin{table*}[h]
    \centering
    \small
    \caption{Performance comparison on \textbf{ScreenSpot-V2}. We use `$^*$' to denote the results evaluated by us. The best results are highlighted in \textbf{bold}, and the second-best results are \underline{underlined}.}
    \label{tab:main_results_screenspot_v2_and_showdown}

    \setlength{\tabcolsep}{3pt} 
    
    \begin{tabularx}{\textwidth}{
        l
        *{6}{>{\centering\arraybackslash}X}
        c
        }
        \toprule
        \multirow{2}{*}{\textbf{Model}} 
        & \multicolumn{2}{c}{\textbf{Mobile}} &
          \multicolumn{2}{c}{\textbf{Desktop}} &
          \multicolumn{2}{c}{\textbf{Web}} &
          \multirow{2}{*}{\textbf{Avg.}}  \\
        \cmidrule(lr){2-3} \cmidrule(lr){4-5} \cmidrule(lr){6-7}
        
        & Text & Icon. & Text & Icon. & Text & Icon. &  \\
        \midrule
        \rowcolor{gray!15}
        \multicolumn{8}{l}{\textit{Open-Source Models}} \\
        Qwen3-VL-2B$^*$~\citep{Qwen3-VL} &95.5 &82.0 &95.4 &73.6 &89.7 &76.4 &\cellcolor{light_purple}86.7 \\
        Phi-ground~\citep{phi_ground}
            & 90.2 & 76.4 & 93.6 & 75.9 & 96.5 & 62.0
            & \cellcolor{light_purple}83.8
            \\
        OS-Atlas-7B~\citep{osaltas_and_screenspot_v2}
            & 95.2 & 75.8 & 90.7 & 63.6 & 90.6 & 77.3
            & \cellcolor{light_purple}85.1
           \\
        UGround-v1-7B~\citep{uground}
            & 83.6 & 90.5 & 85.8 & 86.3 & 95.5 & 83.2
            & \cellcolor{light_purple}87.7
            \\
        UI-Tars-1.5-7B~\citep{ui-tars-15}
            & 92.2 & 81.5 & 91.0 & 84.2 & 95.5 & 84.5
            & \cellcolor{light_purple}89.0
            \\

        SE-GUI-7B$^*$~\citep{SE_GUI}
            & \underline{99.3} & 89.1 &
              96.4 & 78.6 &
              92.7 & 81.3 &
              \cellcolor{light_purple}90.8 
              \\
        UI-TARS-7B~\citep{ui-tars}
            & 96.9 & 89.1 & 95.4 & 85.0 & 93.6 & 85.2
            & \cellcolor{light_purple}91.6
             \\
        Qwen3-VL-8B$^*$~\citep{Qwen3-VL} &97.9 &84.8 &95.9 &87.9 &95.7 &83.7 &\cellcolor{light_purple}91.7\\
        GUI-Actor-7B$^*$~\citep{gui_actor}
            & 97.6 & 88.2 & 96.9 & 85.7 & 93.2 & 86.7
            & \cellcolor{light_purple}92.1
             \\
        OpenCUA-7B~\citep{opencua}
            & - & - & - & - & - & -
            & \cellcolor{light_purple}92.3
            \\
        GTA1-7B~\citep{GTA1}
            & 99.0 & 88.6 & 94.9 & 89.3 & 92.3 & 86.7
            & \cellcolor{light_purple}92.4
             \\
        GUI-Owl-7B~\citep{mobileagentv3}
            & 99.0 & 92.4 & 96.9 & 85.0 & 93.6 & 85.2 & 
            \cellcolor{light_purple}92.8 \\
        GUI-G$^2$-7B$^*$~\citep{GUI_G2}
            & 98.3 & 91.9 &
              95.4 & 89.3 &
              94.0 & 87.7 &
              \cellcolor{light_purple}93.3 
              \\
        InfiGUI-G1-7B$^*$~\citep{infiguig1}
            & 99.0 & 91.9 &
              94.3 & 82.1 &
              \textbf{97.9} & 89.2 &
              \cellcolor{light_purple}93.5 
              \\
        UI-Venus-7B~\citep{ui_venus}
            & 99.0 & 90.0 &
              96.9 & 90.7 &
              96.2 & 88.7 &
              \cellcolor{light_purple}94.1
               \\
        \midrule


        Qwen3-VL-32B$^*$~\citep{Qwen3-VL} & 96.2 &90.0 &97.4 &85.0 &95.7 &89.7 &\cellcolor{light_purple}93.0\\
        GUI-Owl-32B~\citep{mobileagentv3}
            & 98.6 & 90.0 & 97.9 & 87.8 & 94.4 & 86.7 & 
            \cellcolor{light_purple}93.2 \\
        OpenCUA-32B~\citep{opencua}
            & - & - & - & - & - & -
            & \cellcolor{light_purple}93.4
            \\

        GTA1-32B~\citep{GTA1}
            & 99.7 & 90.5 & 99.0 & 94.3 & 95.7 & 90.1
            & \cellcolor{light_purple}95.2
            \\
        UI-Venus-72B~\citep{ui_venus}
            & \textbf{99.7} & \textbf{93.8} & 95.9 & 90.0 & 96.2 & \underline{92.6} & 
            \cellcolor{light_purple}\underline{95.3} \\

        \midrule
        \rowcolor{gray!15}
        \multicolumn{8}{l}{\textit{Ours}} \\
        \textbf{\modelname-2B}& \underline{99.3} &87.2 &97.4 &88.6 &94.0 &84.7 &\cellcolor[HTML]{cfcdfd}92.5\\
        
        \textbf{\modelname-8B}&\underline{99.3} &89.1 &\underline{99.0} &\underline{92.1} &\textbf{97.9} &91.1 &\cellcolor[HTML]{cfcdfd}95.2\\

        \textbf{\modelname-32B}&99.0 &\underline{92.9} &\textbf{99.5} &\textbf{93.6} &\underline{97.4} &\textbf{94.6} &\cellcolor[HTML]{cfcdfd}\textbf{96.5}\\
        \bottomrule
    \end{tabularx}
\end{table*}
\begin{table}[h]
    \caption{Performance comparison of state-of-the-art models on the OSWorld-G-Refine. The best results are highlighted in \textbf{bold}, and the second-best results are \underline{underlined}.}
    \label{tab:osworld_g_refine_comparison}
    \centering
    \setlength{\tabcolsep}{4pt}
    \small
    \begin{tabular}{lcccccc}
        \toprule
        \textbf{Agent Model} & \makecell[c]{\textbf{Text}\\\textbf{Matching}} & \makecell[c]{\textbf{Element}\\\textbf{Recognition}} & \makecell[c]{\textbf{Layout}\\\textbf{Understanding}} & \makecell[c]{\textbf{Fine-grained}\\\textbf{Manipulation}}& \textbf{Refusal} & \cellcolor{white}\textbf{Avg} \\
        \midrule
        \rowcolor{gray!15}
        \multicolumn{7}{l}{\textit{Proprietary Models}} \\
        Operator \cite{OpenAICUA} & - & - & - & - & - & 57.8 \\
        \midrule
        \rowcolor{gray!15}
        \multicolumn{7}{l}{\textit{Open-Source Models}} \\
        Qwen3-VL-2B$^{*}$~\citep{Qwen3-VL} &69.3 &60.9 &69.2 &45.0 &- &\cellcolor{light_purple}57.4 \\
        Jedi-3B \cite{jedi} & - & - & - & - & - &\cellcolor{light_purple}61.0 \\

        Jedi-7B \cite{jedi} & - & - & - & - & - & \cellcolor{light_purple}63.8 \\
        UI-TARS-1.5-7B \cite{ui-tars-15} & 52.6 & 75.4 & 72.4 & \underline{66.7} & 0.0 & \cellcolor{light_purple}64.2\\
        Qwen3-VL-8B$^{*}$~\citep{Qwen3-VL} & 73.9 &68.2 &73.1 &54.4 &- &\cellcolor{light_purple}64.4\\
        GTA1-7B~\citep{GTA1} & 63.2 & \underline{82.1} & 74.2  & \textbf{70.5} & 0.0 & \cellcolor{light_purple}67.7 \\
        \midrule
        Qwen2.5-VL-32B  \cite{qwen25vl}& 57.9 &  70.2 & 73.8 & 49.2 & 0.0 & \cellcolor{light_purple}59.6\\
        OpenCUA-32B \cite{opencua} & 63.2 & 79.9 & \textbf{84.9} & 62.1 & 7.4 & \cellcolor{light_purple}70.2 \\
        Qwen3-VL-32B$^{*}$~\citep{Qwen3-VL} & 77.4 &73.6 &76.3 &57.7 &- &\cellcolor{light_purple}69.0\\
        GTA1-32B~\citep{GTA1} & 63.2  &  \textbf{83.6} & \underline{84.4} & \textbf{70.5} & 0.0 & \cellcolor{light_purple}72.2 \\

        \midrule
        \rowcolor{gray!15}
        \multicolumn{7}{l}{\textit{Ours}} \\
        \textbf{\modelname-2B} & 70.9 &69.1 &72.7 &47.7 &- & \cellcolor[HTML]{cfcdfd}63.5\\
        \rowcolor{gray!10}
        \textit{+ Zoom-In} & 71.3 &71.8 &78.3 &51.0 &- & \cellcolor[HTML]{cfcdfd}66.3 \\
        
        \textbf{\modelname-8B} & 77.4 &73.0 &78.3 &55.7 &- & \cellcolor[HTML]{cfcdfd}68.6\\
        \rowcolor{gray!10}
        \textit{+ Zoom-In} & \underline{79.3} &78.8 &84.2 &59.7 &- & \cellcolor[HTML]{cfcdfd}72.9 \\

        \textbf{\modelname-32B} & \textbf{79.7} &79.4 &81.0 &61.7 &- & \cellcolor[HTML]{cfcdfd}\underline{73.9}\\
        \rowcolor{gray!10}
        \textit{+ Zoom-In} & \underline{79.3} &79.7 &84.2 &64.4 &- & \cellcolor[HTML]{cfcdfd}\textbf{75.0}\\
        \bottomrule
    \end{tabular}
\end{table}

\end{document}